\documentclass[pre,twocolumn,twoside,byrevtex,superscriptaddress]{revtex4}

\usepackage[T1]{fontenc}

\usepackage{graphicx,epsfig,verbatim,enumerate}
\usepackage{amssymb,amsmath}
\usepackage{ifthen}
\newboolean{twocolswitch}

\newcommand{\sindex}[1]{}
\newcommand{\nindex}[1]{}

\newcommand{\etal}{\textit{et al.}}
\newcommand{\www}[1]{\url{#1}}

\usepackage{lettrine}

\usepackage{color}

\newcommand{\frequency}{f}

\newcommand{\frequencythreshold}{f_{\textnormal{thr}}}

\newcommand{\frequencyword}{f_{w;y}}
\newcommand{\frequencymedian}[2]{f_{w;#1,#2}^{\textnormal{med}}}
\newcommand{\frequencycut}[2]{f_{w;#1,#2}^{\textnormal{cut}}}

\newcommand{\flux}{\phi}
\newcommand{\fluxup}{\flux_{\textnormal{up}}}
\newcommand{\fluxdown}{\flux_{\textnormal{down}}}

\newcommand{\zipfrank}{r}
\newcommand{\zipfexponent}{\alpha}

\newcommand{\zipfexponentlower}{\alpha}
\newcommand{\zipfexponentupper}{\alpha'}

\newcommand{\fluxexponentlower}{\mu}
\newcommand{\fluxexponentupper}{\mu'}

\newcommand{\fluxrankexponentlower}{\nu}
\newcommand{\fluxrankexponentupper}{\nu'}

\newcommand{\breakpoint}{\textnormal{b}}
\newcommand{\zipfbreakpoint}{\zipfrank_{\breakpoint}}
\newcommand{\frequencythresholdbreakpoint}{f_{\breakpoint}}

\newcommand{\plainlatex}[1]{}

\setboolean{twocolswitch}{true}

\begin{document}

\title{

Is language evolution grinding to a halt? \\
The scaling of lexical turbulence in English fiction suggests it is not.
}

\author{
\firstname{Eitan Adam}
\surname{Pechenick}
}

\email{eitan.pechenick@gmail.com}

\affiliation{
  Computational Story Lab,
  Vermont Complex Systems Center,
  Vermont Advanced Computing Core,
  \& the Department of Mathematics and Statistics,
  University of Vermont,
  Burlington,
  VT, 05401
}

\author{
\firstname{Christopher M.}
\surname{Danforth}
}
\email{chris.danforth@uvm.edu}

\affiliation{
  Computational Story Lab,
  Vermont Complex Systems Center,
  Vermont Advanced Computing Core,
  \& the Department of Mathematics and Statistics,
  University of Vermont,
  Burlington,
  VT, 05401
}

\author{
\firstname{Peter Sheridan}
\surname{Dodds}
}
\email{peter.dodds@uvm.edu}

\affiliation{
  Computational Story Lab,
  Vermont Complex Systems Center,
  Vermont Advanced Computing Core,
  \& the Department of Mathematics and Statistics,
  University of Vermont,
  Burlington,
  VT, 05401
}

\date{\today}

\begin{abstract}
  
Of basic interest is the quantification of the long term growth of a
language's lexicon as it develops to more completely cover
both a culture's communication requirements and knowledge space.
Here, we explore the usage dynamics of words in the English language
as reflected by
the Google Books 2012 English Fiction corpus.
We critique an earlier method that found decreasing birth and
increasing death rates of words over the second half
of the 20th Century,
showing death rates to be strongly affected by the imposed time cutoff
of the arbitrary present and not increasing dramatically.
We provide a robust,
principled approach to examining lexical evolution
by tracking the volume of word flux
across various relative frequency thresholds.
We show that
while the overall statistical structure of the English language 
remains stable over time in terms of its raw Zipf distribution,
we find evidence of an enduring `lexical turbulence':
The flux of words across frequency thresholds from decade to decade
scales superlinearly with word rank and exhibits a scaling break we
connect to that of Zipf's law.
To better understand the changing lexicon,
we examine the contributions to the Jensen-Shannon divergence
of individual words crossing frequency thresholds.
We also find indications that scholarly works about fiction are strongly 
represented in the 2012 English Fiction corpus, and suggest that a future
revision of the corpus should attempt to separate critical works from fiction itself.

\end{abstract}



\maketitle

\section{Introduction}

In studying any entity or system,
a fundamental scientific goal
is the satisfactory characterization of
temporal dynamics, whether empirically observed, simulated, or theoretically predicted.
For language, there are many kinds and scales of temporal dynamics to consider
such as the introduction and usage decline of specific words~\cite{petersen2012a},
the evolution of accents,
the long term development of individual languages~\cite{petersen2012b},
and
the changes in the overall ecology of human languages
which has now moved well into an era of die off~\cite{abrams2003a}.

Here, we are concerned with the dynamics
of the English language's lexicon.
Primarily, we want to know how the usage of words
has changed in time, and how this is reflected in the
English lexicon's
evolution.
This focus leads us to several core questions:
(1) What are the rates at which words are born and at which they die?
(2) How do we reasonably identify word births and deaths in the first place? 
(3) As the English lexicon has expanded, how 
have overall statistical patterns such as Zipf's law~\cite{zipf1949a}
changed, if at all?
We are especially interested with 
revisiting work on word ``birth'' and ``death'' rates
as performed in~\cite{petersen2012a}.
As we will show, the methods employed in~\cite{petersen2012a}
suffer from boundary effects, and we propose and investigate
an alternative approach insensitive to time range choice.
We also investigate lexical changes at a range of usage frequency levels.

We will perform our analyses using
the Google Books corpus~\cite{michel2011quantitative,lin2012syntactic}
whose 
incredible volume generated from an
extensive coverage of all written works
would seemingly make it an ideal candidate for linguistic research.
However, there are two major caveats that limit its potency
and we will lay them out before proceeding.

In previous research~\cite{pechenick2015a}, 
we broadly explored the 
characteristics and dynamics of
the unfiltered English and English Fiction data sets from both 
the 2009 and 2012 versions of the Google Books corpus.
We showed that the  2009 and 2012
unfiltered English 
data sets
and,
surprisingly, the 2009 English Fiction data set,
all become increasingly influenced by scientific texts
throughout the 1900s,
with medical research language being especially prevalent.
We concluded that,
without sophisticated processing or the provision of extensive metadata,
only the 2012 English Fiction data set is suitable for any kind of
analysis and deduction as it stands.

We also described the confounding problem
of the library-like nature of the Google Books corpus.
Each book is, in principle, represented only once
(re-editions are one exception).
Word frequency is thus a deceptive aspect of the Google Books corpus as book
popularity is not encoded in any way.
Word counts are in no way reflective of how often these words are read---as might be informed
by book sales and library borrowing data---much less spoken by
the general public.
Nevertheless, 
the Google Books corpus registers an imprint of a language's lexicon
and remains worthy of study,
as long as we remain mindful of its nature.

\begin{figure}[tbp!]
  \centering
  \includegraphics[width=\columnwidth]{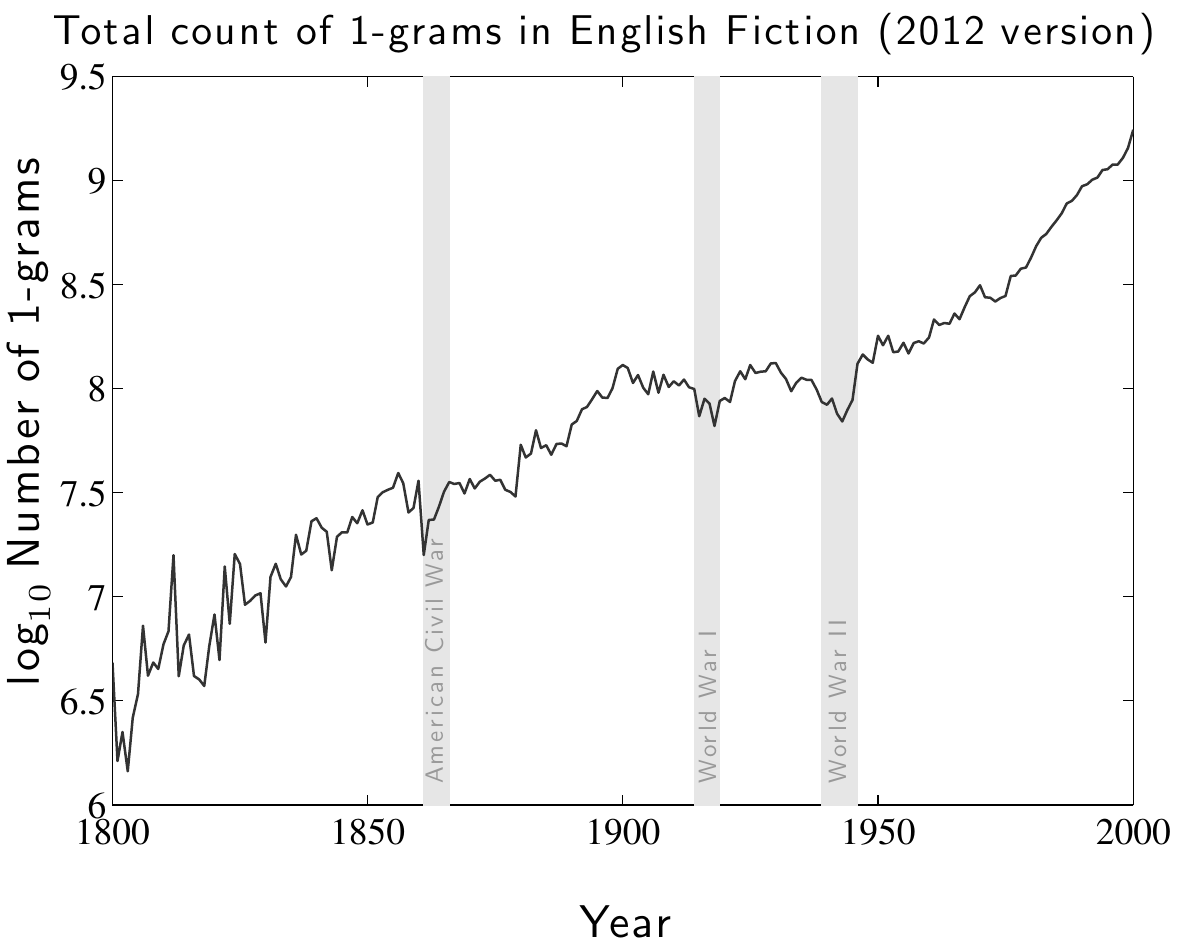}
  \caption{ 
    Total 1-gram counts for the Google
    Books corpus 2012 English Fiction data set
    as a function of publication year.
    More than simple words,
    1-grams include any sequence of unbroken non-space characters
    as well as
    punctuation marks, such as commas and brackets, which are broken away from words.
    The number of 1-grams is proportional to the number of books and pages in the corpus
    but does not account for readership popularity~\cite{pechenick2015a}.
    A roughly exponential
    increase in 1-gram volume is apparent over time with several periods of stasis
    and decline.
    Volume decreases are particularly apparent during
    the American Civil War, World War I, and World War II
    as indicated by the three vertical gray shadings.
    Both World Wars follow a decade of stagnancy and
    decline for the corpus.
  }
\label{fig:langevo.ficvolume}
\end{figure}

In this paper, 
we therefore focus only on the 2012 version of the English Fiction data
set.
To provide a sense of scale for this corpus,
we show in Fig.~\ref{fig:langevo.ficvolume}
the total number of 1-grams 
for this data set between 1800 and 2000
(1-grams are defined to be contiguous text elements and are more general
than words including, for example, punctuation;
for ease of expression,
we will use word and 1-gram
interchangeably).
An exponential increase in volume is
apparent over time with notable exceptions during major conflicts when
the total volume decreases.
There is effectively zero growth in volume over first half of the 20th Century.

A number of researchers have carried out studies of the Google Books
corpus with the aim of examining properties and dynamics of entire languages.
These include analyses of Zipf's and Heaps' laws as applied to the
corpus~\cite{gerlach2013stochastic}, 
the rates of verb
regularization~\cite{michel2011quantitative}, 
rates of word ``birth''
and ``death'' and durations of cultural
memory~\cite{petersen2012a}, 
as well as an observed decrease
in the need for new words in several
languages~\cite{petersen2012b}.
However, most of the studies were performed before the release of the
second version, and, to our knowledge, none have taken into account the substantial
effects of scientific literature on the data sets.

We structure the paper as follows.
In
Sec.~\ref{sec:langevo.critique}, we critique the method
from~\cite{petersen2012a} 
which examines the birth and death
rates of 1-grams for several languages using the first Google Books corpus.
Through a number of different analyses, we show that while 1-gram
birth rate has slowed, death rates have not increased substantially.
In Sec.~\ref{sec:langevo.evolution}, we describe
information theoretic methods for
examining lexical evolution using the Jensen-Shannon divergence,
and then present our observations in the form of word shift graphs.
We first recall
and confirm an apparent bias toward increased usage rates of
1-grams over time~\cite{pechenick2015a}.
We then measure the flux of 1-grams
across various relative frequency boundaries in both directions
for the 2012 English Fiction data set.
We describe the use
of the largest contributions to the Jensen-Shannon divergence between
successive decades from among the 1-grams crossing each boundary as
signals to highlight the specific dynamics of 1-gram growth and decay
over time.
We display ranked examples of
these 1-gram usage changes and explore the factors contributing to the observed
disparities between growth and decay.
In releasing the original data set, Michel~et~al.~\cite{michel2011quantitative}
noted that English Fiction contained scholarly articles about
fictional works (but not scholarly works in general),
and we also investigate this mixing of texts.
We offer concluding remarks in Sec.~\ref{sec:langevo.conc}
Supporting material can be found at our paper's 
Online Appendices~\cite{pechenick2015b_onlineappendices}.

To compare across years, we will work with
\textit{relative frequency},
$f$.
For a 1-gram $w$,
$f_{w;y}$
is
the usage frequency of $w$ normalized by the total
number of 1-grams in the year $y$
(the total number of
1-grams is to be distinguished
form the number of unique
1-grams).
We will also use the average relative frequency
for a 1-gram over the time scale of a decade.

\section{On quantifying the birth and death of words}
\label{sec:langevo.critique}

In this section, we aim to measure the births and
deaths of words over time.
As we will show, this will turn out to be a delicate
and arguably ill-defined task.
We will arrive at this conclusion
by considering and attempting to reproduce the work of
Petersen \etal~\cite{petersen2012a} on word
life spans,
and,
by doing so,
show how word death rates are strongly
affected by our vantage point in history.

Petersen \etal~\cite{petersen2012a} examined
the birth and death rates of words over time for various
data sets in the 2009 version of the Google Books corpus
including unfiltered English, English Fiction,
Spanish, and Hebrew.

Their quantification of birth and death is nuanced
and requires some examination.
They define the birth year and death year of an individual word as
the first and last year, respectively, that the given word's relative
frequency $\frequencyword$ is found to be equal to or greater than
a cutoff frequency $\frequencycut{y_1}{y_2}$ equal to
one twentieth its median relative frequency $\frequencymedian{y_1}{y_2}$,
i.e.,
$
\frequencyword
\ge 
\frequencycut{y_1}{y_2}
=
0.05
\frequencymedian{y_1}{y_2}.
$
The subscripts $y_1$ and $y_2$ indicate the first and last year of the overall time
period.
They exclude words appearing in only one year (we will show this is problematic)
and words appearing for the first time before $y_1 = 1700$. 
The rates of word birth and death, respectively, are then found by
normalizing the numbers of word births and deaths by the total
number of unique words in a given year.

For all four data sets, Petersen \etal\ found
strongly decreased birth rates and
increased death rates over time,
a variation for both of two to three orders of magnitude
occurring most rapidly between 1950 and 2000.
They noted that they obtained qualitatively similar results
when one tenth the median frequency is used as the cutoff threshold.

The very specific nature of the
analysis raise questions as to the robustness of the method.
Three major concerns:
\begin{enumerate}
\item
  Use of the median relative frequency for a threshold of birth and death.
  This quantity depends on the word and the year range chosen.
  Hypothetically, a rare word $w_1$ that has a constant relative frequency over time will
  never be identified as being born or dying,
  while a word $w_2$ with
  much higher relative frequencies may die out yet still
  never fall to the relative frequency of $w_1$.
  In short, the standards for a word's birth and death
  vary from word to word and from
  time range to time range.
  As we will see in particular,
  shifting the end year $y_2$ using this analysis strongly affects death rates.
\item
  The problematic use of the median for very rare words.
  Rare words that have a zero relative frequency
  in more than half of the years examined 
  will have a median relative frequency
  $\frequencymedian{y_1}{y_2}$ = 0.
  The strict definition
  in~\cite{%
    petersen2012a%
  }
  means such a word will never be born or die
  as its relative frequency will always be greater than or equal
  to 0
  (such a state could be possibly termed `unalive'~\cite{pratchett1997b}).
  If we simply ignore such words, then we will at the same time
  be including words with lower overall abundance, e.g., words
  that appear only in two consecutive years.
  To overcome this issue, we adjust Petersen \etal's criterion
  to involve an inequality:
  $
  \frequencyword
  >
  \frequencycut{y_1}{y_2}
  =
  0.05
  \frequencymedian{y_1}{y_2}.
  $
  We note that we presume the computation of the median
  in~\cite{petersen2012a} was carried
  out for the range of years covering the first
  and last appearance of a word.
\item
  By necessity, the Google Books corpus was constructed
  with a frequency threshold for a word to be included
  or not (a word must appear at least 200 times).
  Thus, a word having a relative frequency of 0 in the data set
  does not mean it was entirely absent.
  We do not attempt to incorporate this issue of censusing
  here but note that it becomes problematic for rare words
  (which are collectively legion).
\end{enumerate}

\begin{figure*}[tbp!]
  \centering
  \includegraphics[width=\textwidth]{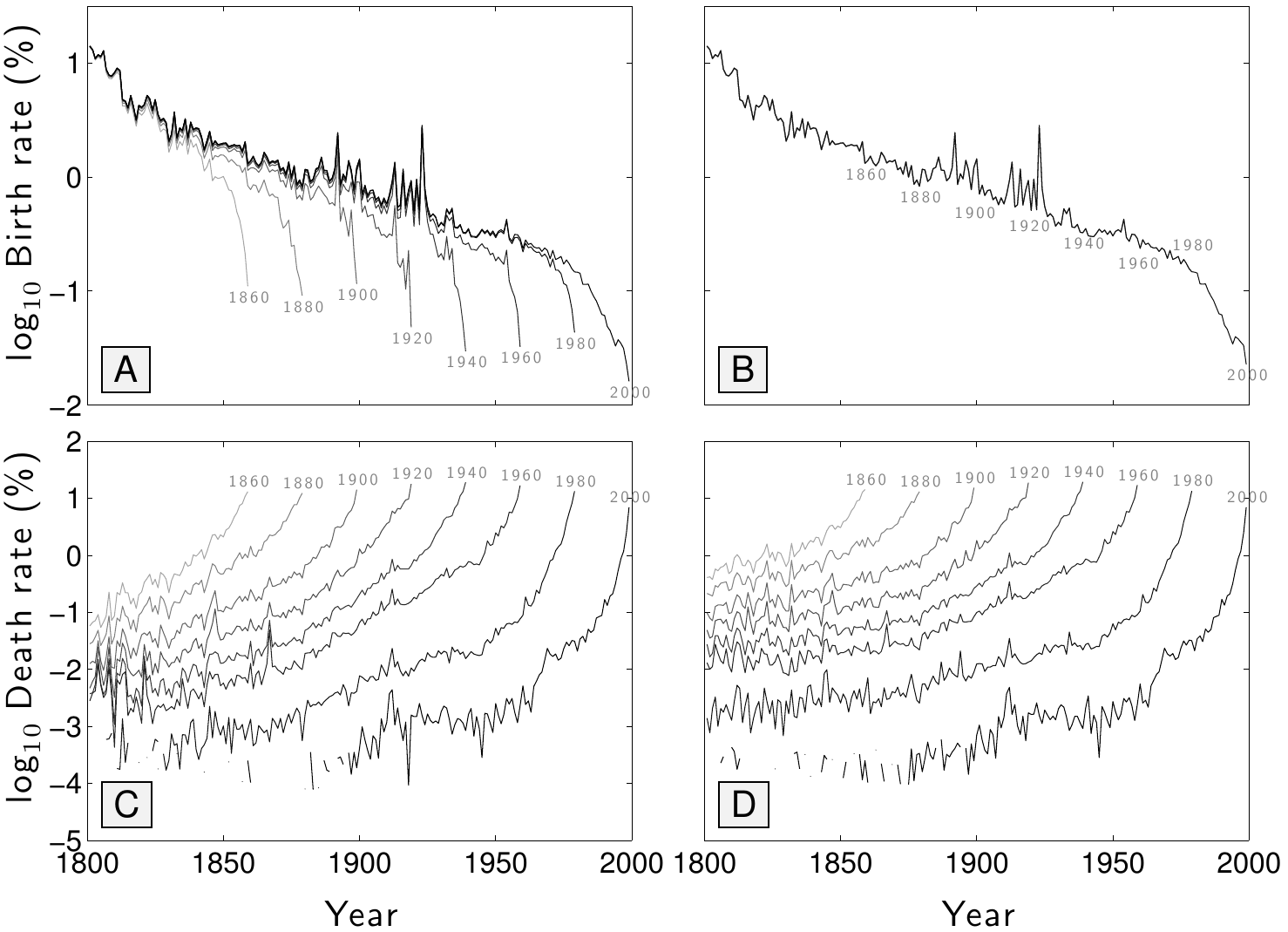}
  \caption{
    \textbf{A} and \textbf{C}:
    Birth and death rates for 1-grams
    for the 2012 version of English Fiction
    determined using the method 
    proposed in~\protect\cite{petersen2012a}.
    Curves correspond to different end-of-history boundaries
    with history running from $y_1$=1800 to $y_2$=1860 to 2000
    in 20 year increments.
    Birth rates show clear departures from an overall form
    as each end of history year is approached.
    Including words that appear in only one year in a time
    range eliminates these discrepancies (plot \textbf{B}).
    Death rates however are strongly affected by the choice
    of when history ends and this cannot be remedied by modifying
    the rule for 1-gram death.
    As the end of history moves forward in time,
    words that seemed dead are no longer dead
    for a number of reasons
    \textbf{B} and \textbf{D}:
    Birth and death rates as per plots \textbf{A} and \textbf{C} in all respects
    except now including words that appear only once in a time range---i.e., have a non-zero relative
    frequency in only one year.
    Birth rates are now well determined retrospectively from any vantage point of history
    and an exponential decay appears confirmed.
    Death rates remain incongruent as
    in \textbf{C}.
  }
  \label{fig:langevo.petersen1}
\end{figure*}

\begin{figure*}[tbp!]
  \centering
            \includegraphics[width=\textwidth]{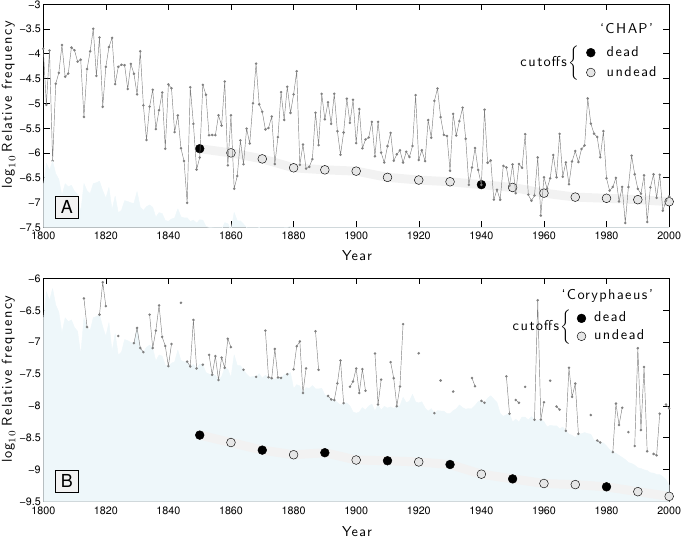}
  \caption{
    Two examples of how a 1-gram may be variously labeled dead or alive depending
    on the end of history using the criterion in~\cite{petersen2012a}.
    \textbf{A.}
    The word `CHAP' declines in relative frequency over time, from
    a high of $10^{-3.5}$ to as low as $10^{-7.5}$.
    Using a twentieth of the median frequency of a 1-gram
    as a threshold for birth and death, we see `CHAP' appears
    to have ``run down the curtain'' in 1850 but then re-emerged
    as alive for 8 subsequent decadal end points.
    `CHAP' once again succumbs in 1940 only to
    stagger on through 2000.
    This dead-undead cycling can be seen for many words 
    and leads us to exploring how words pass above
    and drop below fixed relative frequency thresholds.
    In both plots, the blue region marks the lowest possible
    relative frequency for each year achieved when a 1-gram has a count of 1.
    \textbf{B.}
    The word `Coryphaeus' is a much less frequent word than `CHAP',
    and its time series contains a substantial number of zeroes
    and ones (resting on the top of the blue region).
    The criterion in~\cite{petersen2012a} leads to
    a flipping back and forth between being dead and undead at most end-of-history
    years from 1850 through to 2000.
  }
  \label{fig:langevo.examplepetersenfail}
\end{figure*}

With these points in mind,
we recreated the described analysis of~\cite{petersen2012a}
for the 2012 version of English Fiction at the level of individual years.
Per~\cite{petersen2012a}, 
we initially exclude words appearing in only one year.
We also limit our analysis to 1800-- (rather than 1700--)
and our findings will help us address this choice.
We believe these differences with~\cite{petersen2012a} 
should not be substantive, and allow us to re-examine their
work and build out our own in meaningful ways.

We compare the birth and death rates as observed at
different end points of history
by performing the analysis with
$y_2$=1860 through to $y_2$=2000 in increments of 20 years.
We present the resulting birth and death rates in
Fig.~\ref{fig:langevo.petersen1}
(cf. Fig.~2 in~\cite{petersen2012a}).
Subsequent modifications which we will explain below
will give us the comparison plots
shown in Figs.~\ref{fig:langevo.petersen1}B and D.

We observe
in Fig.~\ref{fig:langevo.petersen1}A
that birth rates decline approximately exponentially overall,
and here we find general agreement with~\cite{petersen2012a}
and~\cite{petersen2012b}.
However, we also see sharp departures to much lower birth rates
near the end point of each history.
We are able to entirely attribute these drops to
the decision in~\cite{petersen2012a}
to ignore all
1-grams that have a non-zero relative
frequency in a single year.
By including these 1-grams,
Fig.~\ref{fig:langevo.petersen1}A
becomes
Fig.~\ref{fig:langevo.petersen1}B
and we see that birth rate is no longer
affected by the choice of $y_2$.
We therefore see that words that appeared in only one year before
a selected end year $y_2$ may well be just sputtering into
existence.
Such words will be retrospectively declared born when
the end of history moves forward.

We can also now see that the apparent speeding
up of the drop in birth rate after 1980
in Fig.~\ref{fig:langevo.petersen1}B
appears to be real, consistent with~\cite{petersen2012a}.
We note that this is a complicated time
with massive growth and change in information technology and publishing,
and we will see that literary criticism starts to
populate the corpus during this time period as well.

We now turn to word death rates in Fig.~\ref{fig:langevo.petersen1}B.
In contrast to birth rates, there is no overlap between
death rates at any point in time as a function of the end of history $y_2$.
For example, death rates in the late 1800s are estimated at 10\% if
$y_2$=1900 but $< 10^{-3}$\,\% if $y_2=2000$.
It appears that words are not in fact dying out.

So why is the word death rate used in~\cite{petersen2012a}
affected so profoundly by boundaries?
Including words appearing in only one year as we did for birth rates,
does not resolve this issue:
Fig.~\ref{fig:langevo.petersen1}D
is essentially the same as
Fig.~\ref{fig:langevo.petersen1}C.

The problem lies instead
in that the relative frequency threshold for a word ``existing'' in a given
year $y$ is determined by range of years being considered.
We argue that a number of example relative frequency trajectories
are problematic for a range dependent definition.

Consider two different ranges of years,
$[y_1,y_2]$
and
$[y_1,y'_2]$
with 
$y_2 < y'_2$
and a year $y$
internal to both ranges.
The median relative frequency 
for the same 1-gram will very likely differ
for the two ranges,
and a word which is alive in year $y$ for the $[y_1,y_2]$ range
may be either not yet born
or dead for the same year $y$
in the $[y'_1,y'_2]$ range.

This complication allows for unintuitive results
such as a 1-gram appearing to have died out
by $y_2$ but over a longer period of time
ending at $y'_2$, it qualifies as having
being alive, or possibly, ``undead.''

We provide two examples of dead-undead behavior
in Fig.~\ref{fig:langevo.examplepetersenfail}.
First,
in Fig.~\ref{fig:langevo.examplepetersenfail}A,
we show the word ``CHAP'' (all capitals, likely short for Chapter).
We chose this word as one with a reasonably
high median relative frequency but otherwise
fairly randomly from all words with
oscillating dead-undead states.
The main curve is the relative frequency
for `CHAP' over time showing a gradual
decline over around three orders of
magnitude.
In both Figs.~\ref{fig:langevo.examplepetersenfail}A
and~\ref{fig:langevo.examplepetersenfail}B,
the blue region outlines the lowest possible relative
frequency for each year (i.e., 1 divided by the total number of 1-grams recorded).

We measure median relative frequency
over a series of time ranges
with $y_1$=1800 and ends-of-history
at $y_2$=1850 through to $y_2$=2000 in decade steps.
The circles mark the cutoff frequency $\frequencycut{y_1}{y_2}$
for each time range.
Open circles indicate the relative
frequency of `CHAP' has exceeded the cutoff at
that $y_2$---`CHAP' is alive---while
filled circles show that `CHAP' has died.

In 1850, the word `CHAP' would have 
appeared to have snuffed it in 1848;
then viewed as having only temporarily been stunned
and revived for the following 8 decadal end points;
been declared an ex-word again in 1940,
nailed to the perch as it were;
and finally
seen again to be only resting
and not at all ready to
push up the daises through to 2000~\cite{montypython1969deadparrotsketch}.

In Fig.~\ref{fig:langevo.examplepetersenfail}B,
we show the relative frequency for a much less common word which displays
a different kind of dead-undead cycling: ``Coryphaeus'' (the head of a Greek chorus).
The time series includes numerous zeroes (which we must remember pertain only to
the sample behind the Google Books 2012 English Fiction corpus).
This example shows a decadal-scale swapping between being dead and undead from 1850 on,
and demonstrates how zero frequencies may induce unexpected
behavior in the birth-death criterion in~\cite{petersen2012a}.
Essentially, whenever ``Coryphaeus'' does not appear in the corpus for a year,
it will be considered dead, and if it does appear, it will have a relative
frequency exceeding the dead-undead cutoff.

Thus, while the method in~\cite{petersen2012a}
provides a reasonable approach to analyzing
dynamics and asymmetries in the evolutionary dynamics of a language
data set and is informative about birth rates,
the results for death rates 
depend on when the experiment is performed.
We proceed to develop an approach that is independent of time boundaries
and agnostic to the 1-grams themselves.

\section{Tracking language evolution through the flux of words across relative frequency thresholds}
\label{sec:langevo.evolution}

We move away from attempting to identify
words as having been born or died to exploring the flux of words
across fixed relative frequency thresholds.
For example, over some time span,
we wish to find and count which words decline in prevalence and drop below,
say, a relative frequency of $10^{-5}$, along with which words move up above
this threshold.
With a decay in the birth rates of words,
English may be globally ``cooling''~\cite{petersen2012b}
but we will show that there is still much bubbling within.

To work at a meaningful temporal scale,
we coarse-grain the relative frequencies in the second English Fiction
data set at the level of decades---e.g., between 1870-to-1879 and
1990-to-1999---by averaging the relative frequency of each unique word
in a given decade over all years in that decade.
We weight each year
equally.
This allows us to conveniently calculate and sort
contributions to the Jensen-Shannon divergence (defined below)
of individual 1-grams between, for example, any two time periods.
To avoid high levels of optical character recognition (OCR) errors
for texts typeset prior to the early 19th century, 
we will concern ourselves going forward with 1-grams between the years 1820 and 2000. 
A prevalent example is the long s---e.g., ``said'' being read as
``faid''~\cite{pechenick2015a}.

\subsection{Basic stability of Zipf's law}
\label{langevo:subsec.zipfslaw}

\begin{figure*}[tbp!]
  \centering
  \includegraphics[width=\textwidth]{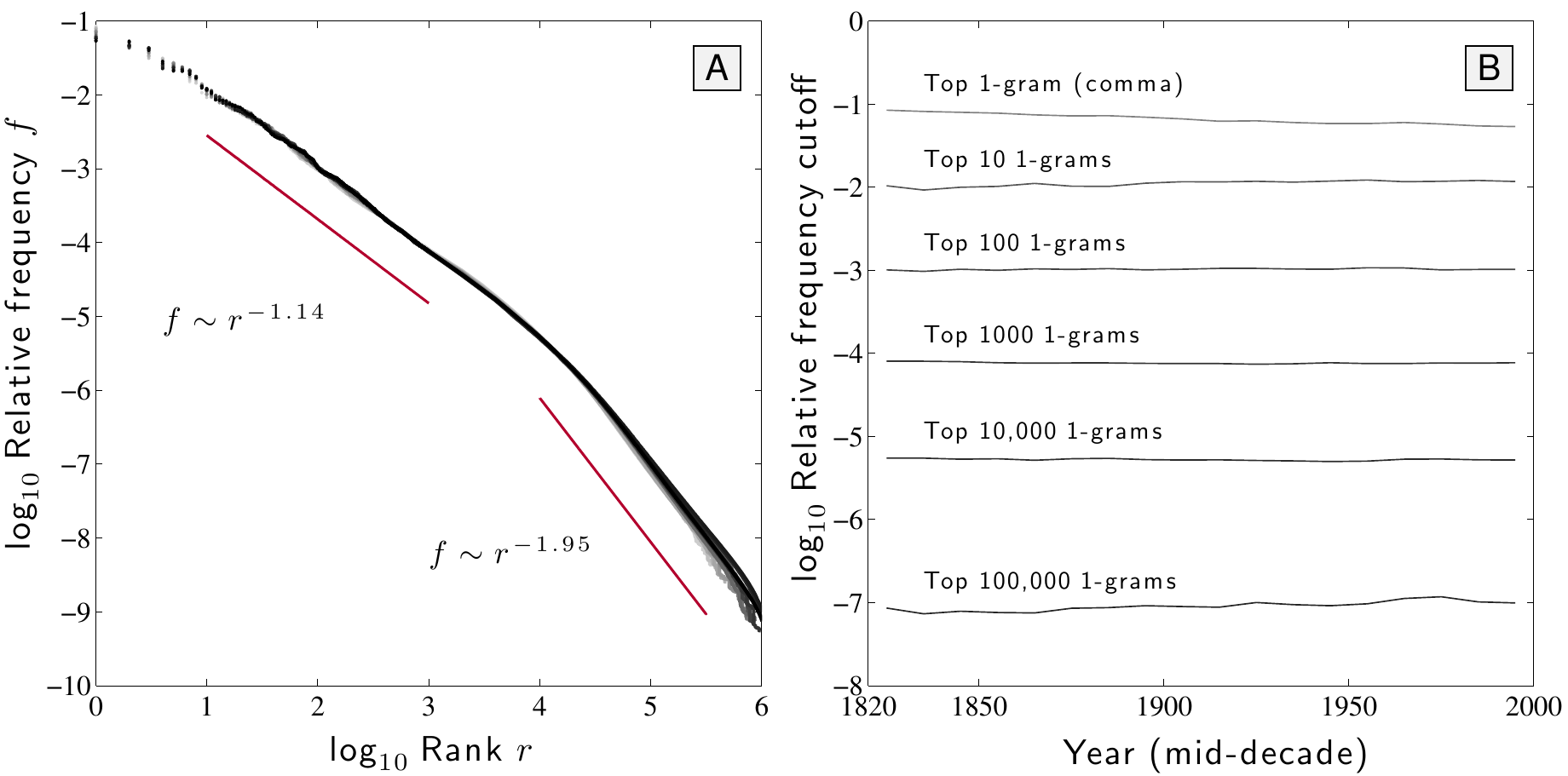}
  \caption{
    \textbf{A.}
    Overlay of Zipf's laws for all decades running
    from the 1820s to the 1990s for the 2012 English
    Fiction corpus (color shifts from light gray to black).
    The observed stability demonstrates that Zipf's law remains largely
    unchanged for the 2012 English Fiction data set, even though
    individual words may vary greatly in rank over time.
    \textbf{B.}
    In support of \textbf{A},
    rank threshold boundaries correspond to nearly constant
    relative frequency threshold boundaries over many orders of
    magnitude, with the exception of the top 1-gram (always a comma),
    which decreases in relative frequency.    
    Points are located at the center of each decade.
  }
  \label{fig:langevo.zipfslaw}
\end{figure*}

A famous and fundamental scaling for language is
Zipf's law~\cite{zipf1949a} which
was long held to be that the relative frequency of a word in a corpus
scales approximately as the inverse of its size rank,
$\frequency \sim \zipfrank^{-\zipfexponent}$ 
with $\zipfexponent \simeq 1$.
However, recent empirical work has shown that for large corpora,
Zipf's law typically exhibits two scaling regimes:
\begin{equation}
  \frequency
  \sim
  \left\{
  \begin{array}{c}
    \zipfrank^{-\zipfexponentlower} \ \mbox{for}\ \zipfrank \ll \zipfbreakpoint,\\
    \zipfrank^{-\zipfexponentupper} \ \mbox{for}\ \zipfrank \gg \zipfbreakpoint,
  \end{array}
  \right.
  \label{langevo.eq:zipfslawtwoscaling}
\end{equation}
where
$\zipfexponentupper > \zipfexponentlower$,
and
the transition between scaling regimes
around the break point $\zipfbreakpoint$
typically occurs over an order of magnitude.
Prior work by our group has elsewhere found the break in scaling for Zipf's law
to be a result of text mixing~\cite{williams2015b}
(other theories have been put forward~\cite{ferrericancho2001c,gerlach2013a}).
The break point $\zipfbreakpoint$ can be estimated by average text length,
though we cannot do so for the Google Books
corpus as the necessary information on individual books is not available.

For the present work, we only need
to characterize Zipf's law with its two scaling regimes.
In Fig.~\ref{fig:langevo.zipfslaw}A,
we plot Zipf's law for each decade running from the 1830s through to the 1990s.
We observe very strong agreement over nearly 200 years
of English Fiction.
The variations that we do see are
(1) the most common words become slightly less common,
and
(2) the tail becomes slightly fatter as new 1-grams enter the lexicon.

For the sake of introducing and broadly characterizing word flux,
it is sufficient for us to perform a simple measure of the scaling exponents
by averaging the Zipf's laws and then using standard
linear regression over
the ranges indicated in Fig.~\ref{fig:langevo.zipfslaw}A.
We estimate
$\zipfexponentupper \simeq 1.14$
and
$\zipfexponentlower \simeq 1.95$.

In Fig.~\ref{fig:langevo.zipfslaw}B,
we show in detail how the numbers of 1-grams with relative frequencies
exceeding fixed thresholds are stable over time.
The only exception is the top 1-gram---always the comma---which gradually deflates in relative frequency
(punctured punctuation).

At least in the case of English fiction then,
the ``bones'' of Zipf's law have changed little over
the period 1820 to 2000.

But the words underlying Zipf's law have fluctuated in relative frequency,
and this is an aspect often overlooked when comparing ranked  distributions
for any system.

\subsection{Lexical turbulence: The scaling of word flux across internal frequency thresholds}
\label{langevo:subsec.wordfluxscaling}

\begin{figure*}[tbp!]
  \centering
  \includegraphics[width=\textwidth]{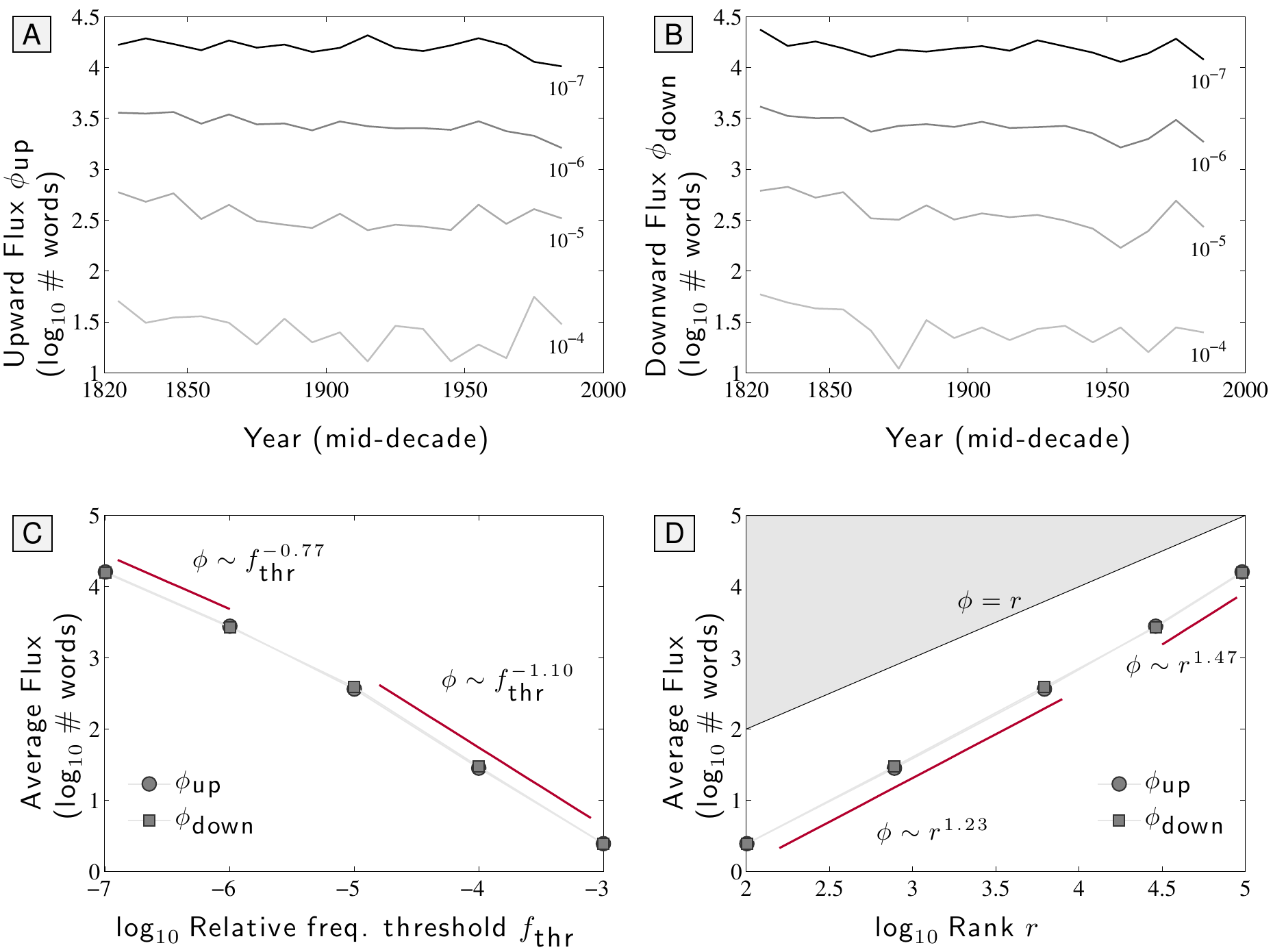}
  \caption{
    Upward and downward fluxes,
    $\fluxup$
    and
    $\fluxdown$
    as a function of relative frequency threshold
    $\frequencythreshold$.
    These fluxes are the total number of words crossing 
    relative frequency thresholds of
    $\frequencythreshold=10^{-4}$,
    $10^{-5}$,
    $10^{-6}$,
    and
    $10^{-7}$
    in both
    upward (Plot \textbf{A})
    and
    downward (Plot \textbf{B})
    directions between consecutive decades.
    Word flux is roughly proportional to the inverse
    of the relative frequency threshold $\frequencythreshold$.
    For each threshold, the upward and downward flux roughly
    cancel.
    For either direction of flux, there appears to be little
    qualitative difference between the three smallest thresholds.
    Plot \textbf{C}:
    Upward and downward fluxes
    $\fluxup$
    and
    $\fluxdown$
    from
    \textbf{A} and \textbf{B}
    averaged over all decade pairs
    as a function of relative frequency threshold
    $\frequencythreshold$.
    The balance between the two fluxes is evident,
    and the two scaling regimes identified 
    are tied to the break in scaling in Zipf's law
    (Fig.~\ref{fig:langevo.zipfslaw}A)~\cite{williams2015b}.
    Word flux scales approximately as the inverse
    of frequency threshold showing that in moving
    further away from the most common words,
    the English language becomes more and more alive,
    churning internally in
    a regular fashion.
    Plot \textbf{D}:
    Using Zipf's law from Fig.~\ref{fig:langevo.zipfslaw}A,
    \textbf{C} transformed to show flux as a function
    of word rank.
    The scaling regimes connect to those of
    \textbf{C}
    through the Zipf exponents
    $\zipfexponentlower$
    and
    $\zipfexponentupper$
    (See Eqs.~\ref{langevo.eq:zipfslawtwoscaling},
    \ref{langevo.eq:fluxscaling1},
    and
    \ref{langevo.eq:fluxscaling2}).
    The superlinear scaling makes clear
    that the growth in lexical turbulence with rank $\zipfrank$
    is strong.
    For example, around 16.8\% of words with $\zipfrank \le 10^5$
    will be replaced every decade.
  }
  \label{fig:langevo.threshold_crossings}
\end{figure*}

In Fig.~\ref{fig:langevo.threshold_crossings},
we show word flux as a function of time and frequency threshold.
First,
In Figs.~\ref{fig:langevo.threshold_crossings}A
and~\ref{fig:langevo.threshold_crossings}B,
we display the upward and downward fluxes
$\fluxup$
and
$\fluxdown$
of the number of 1-grams
crossing relative frequency thresholds of powers of
10 from
$\frequencythreshold=10^{-4}$ down to $10^{-7}$.
Each point is centered in a decade and represents the total number
of words moving across a frequency threshold 
from that decade to the next.

We can see that word flux across frequency thresholds is
relatively constant over time.  Of the minor modulations
we see some consistency across thresholds, notably recent
decades for $\fluxdown$
and
$\frequencythreshold=10^{-5}$,
$10^{-6}$,
and
$10^{-7}$.
Moreover,
in comparing
Figs.~\ref{fig:langevo.threshold_crossings}A
and~\ref{fig:langevo.threshold_crossings}B,
the two fluxes appear
to be fairly balanced.

However, word flux does vary strongly
with respect to frequency threshold $\frequencythreshold$
and we view this as a kind of `lexical turbulence'.
We see in
Figs.~\ref{fig:langevo.threshold_crossings}A
and~\ref{fig:langevo.threshold_crossings}B
that, as we should expect, the lower the threshold, the higher the flux.
The most common words have essentially no turnover (see below)
while increasingly rare ones are increasingly volatile.

In Fig.~\ref{fig:langevo.threshold_crossings}C,
we attempt to characterize the relationship between
word flux and frequency threshold,
$\frequencythreshold$.
We average the fluxes in
Figs.~\ref{fig:langevo.threshold_crossings}A
and~\ref{fig:langevo.threshold_crossings}B,
and plot them as a function of
$\frequencythreshold$.
The averages for
$\fluxup$
and
$\fluxdown$
are indistinguishable to the eye,
confirming the balance suggested
in 
Figs.~\ref{fig:langevo.threshold_crossings}A
and~\ref{fig:langevo.threshold_crossings}B.

We have at hand evidence of
lexical turbulence 
through an apparent inverse scaling of word flux across
frequency thresholds,
and we mark two possible scaling regimes:
\begin{equation}
  \flux
  \sim
  \left\{
  \begin{array}{c}
    \frequencythreshold^{-\fluxexponentlower}
    \ \mbox{for}\
    \frequencythreshold \ll \frequencythresholdbreakpoint,
    \\
    \frequencythreshold^{-\fluxexponentupper}
    \ \mbox{for}\
    \frequencythreshold \gg \frequencythresholdbreakpoint,
  \end{array}
  \right.
  \label{langevo.eq:fluxscaling1}
\end{equation}
where
$\fluxexponentlower \simeq 0.77$
and
$\fluxexponentupper \simeq 1.10$,
and
$\frequencythresholdbreakpoint$
is the scaling break point.

In Fig.~\ref{fig:langevo.threshold_crossings}D,
we also show how flux scales with word rank $\zipfrank$.
To do so, we have used the average form of Zipf's law
in Fig.~\ref{fig:langevo.zipfslaw}A
to map frequency to rank.
The evident upper limit for flux
is $\flux = \zipfrank$,
marked by  gray area in Fig.~\ref{fig:langevo.threshold_crossings}D.

We are able to connect the scaling break
for flux with respect to both
$\frequencythreshold$
and
$\zipfrank$
to the scaling break in Zipf's law.
Combining
Eqs.~\ref{langevo.eq:zipfslawtwoscaling}
and~\ref{langevo.eq:fluxscaling1},
we have 
\begin{equation}
  \flux
  \sim
  \left\{
  \begin{array}{c}
    \zipfrank^{\fluxrankexponentlower}
    =
    \zipfrank^{\zipfexponentlower\fluxexponentupper}
    \ \mbox{for}\
    \zipfrank \ll \zipfbreakpoint,
    \\
    \zipfrank^{\fluxrankexponentupper}
    =
    \zipfrank^{\zipfexponentupper\fluxexponentlower}
    \ \mbox{for}\
    \zipfrank \gg \zipfbreakpoint.
  \end{array}
  \right.
  \label{langevo.eq:fluxscaling2}
\end{equation}
We measure the lower and upper exponents in Fig.~\ref{fig:langevo.threshold_crossings}D
as 1.23 and 1.47, and these compare favorably using the equation
above and
the exponents
measured in Figs.~\ref{fig:langevo.zipfslaw}
and~\ref{fig:langevo.threshold_crossings}D:
$
\fluxrankexponentlower
=
\zipfexponentlower\fluxexponentupper
\simeq
1.14
\times
1.10
\simeq
1.25
$
and
$
\fluxrankexponentupper
=
\zipfexponentupper\fluxexponentlower
\simeq
1.95
\times
0.77
\simeq
1.50.
$

Now, both lower and upper scalings of flux with rank
are superlinear meaning that the lexical turbulence
increases strongly with rank---relatively more and more words
are turned over the further we move up in word rank
(down in relative frequency thresholds).
Clearly this scaling cannot be sustained as eventually
we would have $\flux > \zipfrank$.
For the 2012 English Fiction corpus, we see that
the lexicon is exhausted before such a possibility comes about.

For the top 100 words, we see the lexicon is
strongly conserved---crystallized---with on average 2.4\% of words
turning over every decade.
But the superlinear scaling means the lexicon
becomes increasingly volatile.
As we travel out to the top $10^5$ words, the flux has grown
to a considerable 16.8\% per decade.

There are some important limitations to our findings.
The time scale of comparison, which is here
decade-to-decade, will affect the scaling as well,
i.e.,
we need to consider
$
\flux(\zipfrank,\tau)
$
where
$\tau$
is the length in years of adjacent periods.
Clearly, the smaller the time scale of comparison, the less
the degree of lexical turbulence which must tend
toward 0.

We stress that the scalings indicated for flux
are intended only to be rough estimates,
and we will stop well short of proclaiming a set of universal
exponents.
Future work will need to be performed across many languages and
using better curated and different corpora.

In sum, we find that despite a steady decline
in word birth rate for the 2012 English Fiction corpus---two orders of magnitude over two hundred years (Fig.~\ref{fig:langevo.petersen1})---the
flux of words across frequency thresholds
in the Zipf distribution has remained essentially constant in magnitude and scaling.
Our next and last task will be to explore the
individual words most strongly contributing to this lexical turbulence.

\section{Fine-grained exploration of flux across frequency threshold boundaries}
\label{langevo:sec.detailedflux10-2}

We now begin to examine the specific 1-grams that cross relative frequency thresholds
as we move from decade to decade.
We first describe the very limited flux across the
$\frequencythreshold$ = $10^{-2}$
boundary and then
investigate the richer transitions for the lower thresholds
$10^{-3}$ down to $10^{-6}$.

Flux across the $10^{-2}$
boundary between consecutive decades is almost nonexistent from
the 1820s to the 1990s.
The 1-grams that do achieve such a crossing make for a short list of three:
\begin{enumerate}
\item 
  Between the 1820s and 1830s, the semicolon falls
  below the $10^{-2}$ relative frequency threshold.
\item
  Between the 1840s and 1850s, ``I'' rises above
  the $10^{-2}$ relative frequency threshold.
\item 
  Between the 1910s and 1920s, ``was'' rises above
  the $10^{-2}$ relative frequency threshold.
\end{enumerate}
This
is the entirety of the flux across the $10^{-2}$
threshold from 1820 to 2000
showing once again that the regime
of 1-grams above this frequency (roughly the top 10 1-grams) is
extremely 
stable. The eleven 1-grams with relative frequency above
a threshold of $10^{-2}$ in the 1990s in decreasing
order of frequency are:
the comma ``,''
, the period ``.'',
``the'',
the quotation mark ``"'',
``to'',
``and'',
``of'',
``a'', `
`I'',
``in'',
and
``was''.

\subsection{Jensen-Shannon divergence and individual 1-gram contributions}
\label{langevo:subsec.jsddecades}

To enable us to make better sense of
the detailed flux across lower frequency thresholds,
we need some way of assigning some kind of weight of importance to
each 1-gram involved in the flux.
To do so, we start with a standard measure for comparing two probability distributions,
the Jensen-Shannon divergence (JSD)~\cite{lin1991divergence}.
We will then decompose the JSD into contributions from
individual 1-grams which in turn will afford a simple ranking of 1-grams.
We note that other approaches to determining the salience
of words are possible such as the different lens generated
by the use of the  partial KL in~\cite{klingenstein2014a}.

Given two corpora with 1-gram distributions $P$ and $Q$,
the JSD between $P$ and $Q$ may be expressed as
\begin{equation}
  D_{JS}(P\,||\,Q) 
  = 
  H(M) 
  - 
  \frac{1}{2}
  \left[
    H(P)
    +
    H(Q)
    \right],
  \label{langevo:eq.jsd}
\end{equation}
where
$
M=\frac12(P+Q)
$
is the mixed distribution of the two years, and
$
H(P)=-\sum_ip_i\log_2 p_i
$
is the Shannon
entropy~\cite{shannon1948a} of the original distribution.
The JSD is symmetric and bounded between 0 bits and 1 bit,
and these bounds are only attained when the distributions are identical
and free of overlap, respectively.

Helpfully, the JSD is a linear combination of contributions due to individual
words and can be expressed as
$
D_{JS,i}(P\,||\,Q)
=
\sum_{i}
D_{JS,i}(P\,||\,Q).
$
The contribution from the $i^{\textnormal{th}}$ word to the divergence between
the two distributions, as derived from Eq.~\ref{langevo:eq.jsd}, is given by
\begin{equation}
  D_{JS,i}(P\,||\,Q) 
  = 
  m_i
  \cdot
  \frac{1}{2}
  \big[
  r_i
  \log_{2} r_i
  +
  (2-r_i)
  \log_{2} (2-r_i)
  \big],
  \label{langevo:eq.contribution-messy}
\end{equation}
where $r_i=\min(p_i,q_i)/m_i$.
The contribution from an individual word is therefore
proportional to the average frequency of the word $m_i$ and also
depends on the ratio between the smaller and average frequencies, $r_i=p_i/m_i$.
We write the contribution
of the $i^{\textnormal{th}}$ word as:
\begin{equation}
  D_{JS,i}(P\,||\,Q) = m_iC(r_i),
  \label{langevo:eq.contribution}
\end{equation}
where
$
C(r_i)
=
\frac{1}{2}
\big[
r_i
\log_{2} r_i
+
(2-r_i)
\log_{2} (2-r_i)
\big].
$

Words with larger average frequencies ($m_i$)
yield larger contribution
signals as do those with smaller ratios ($r_i$).
A commonly occurring 1-gram changing subtly can produce a large
signal.
So can an uncommon or new word given a sufficient shift
in probability.
The quantity $C(r_i)$ is concave (up) and symmetric about $r_i=1$, where the
frequency remains unchanged
($p_i=q_i=m_i$)
yielding no contribution.
If a word appears or disappears between two decades 
(e.g., $p_i=0$ and $q_i>0$),
then
the contribution is maximized at precisely the average frequency of
the word in question.

\subsection{Asymmetry in Jensen-Shannon divergence measures between decades}
\label{langevo:subsec.JSFasymmetry}

\begin{figure}[tbp!]
  \centering
  \includegraphics[width=\columnwidth]{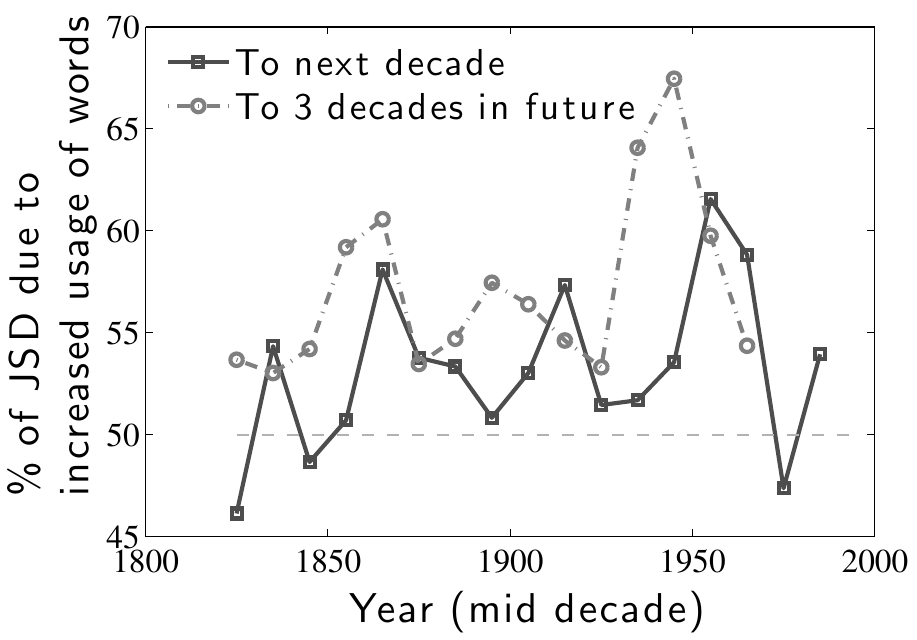}
  \caption{ 
    Percent of (Jensen-Shannon divergence) JSD in English Fiction 2012 corpus due to words
    increasing in relative frequency of use
    for successive decades (dark gray),
    and decades three apart (light gray; e.g., 1990s versus 1960s).
    The contribution for successive
    decades is nearly always more than half---the exceptions are
    between the 1820s, 1840s, and 1970s, and their successive
    decades.
    For decades three apart, the contribution is always
    greater than 50\%.
    The JSD between successive decades
    also shows peaks in the vicinity of major conflicts.
  }
  \label{fig:langevo.jsd_rising_ratios}
\end{figure}

As we show in Fig.~\ref{fig:langevo.jsd_rising_ratios}, more than half
of the JSD between a given decade and the next is typically due to
contributions from words increasing in relative frequency.
The JSDs
between 1820s, 1840s, and 1970s and their successive decades are the
only exceptions.
Moreover, when the time differential is increased to
three decades, no exceptions remain.
This asymmetry is sympathetic to the lexicon enjoying new words
but relatively few true deaths (Sec.~\ref{sec:langevo.critique}).

We note relative extrema of the inter-decade JSD in the vicinity of
major conflicts.
Between the 1860s and successive decade, words on
the rise contribute substantially to the JSD.
This is consistent with
words not relatively popular during wartime (specifically the American
Civil War) being used more frequently in peacetime.
A similar tendency
holds for the JSD between the 1910s (World War I) and the 1920s.
This
is not as apparent in the JSD between the 1910s and the 1940s,
possibly because the 1940s coincide with World War II.
The absolute
maximum for the single-decade curve corresponds to the divergence
between the 1950s and 1960s.
This suggests a strong effect from social
movements.
For the 3-decade split, the absolute peak comes from the
JSD between the 1940s and 1970s, which are certainly decades
of starkly different character.

\subsection{Fine-grained exploration of flux across frequency threshold boundaries: $\frequencythreshold = 10^{-3}$}
\label{langevo:subsec.detailedflux10-3}

We conclude our analysis with a series of
observations on which words contribute
to flux between a number of example
decade pairs and across the frequency
thresholds
$10^{-3}$,
$10^{-4}$,
$10^{-5}$,
and
$10^{-6}$.
For thresholds of $10^{-5}$ and below, we omit signals corresponding to
references to specific years, as such references would otherwise
overwhelm the charts for these thresholds.
We prepare the reader by noting
that these final sections are somewhat detailed in nature.

But we also add that any study of texts reduced to $1$-grams
should in some fashion ``look at the words'' themselves, for the very least as a sanity
check on code and more deeply to find the story behind observed
summarized dynamics and patterns~\cite{dodds2011e,dodds2015a}.

\begin{figure*}[tbp!]
  \centering
  \includegraphics[width=\textwidth]{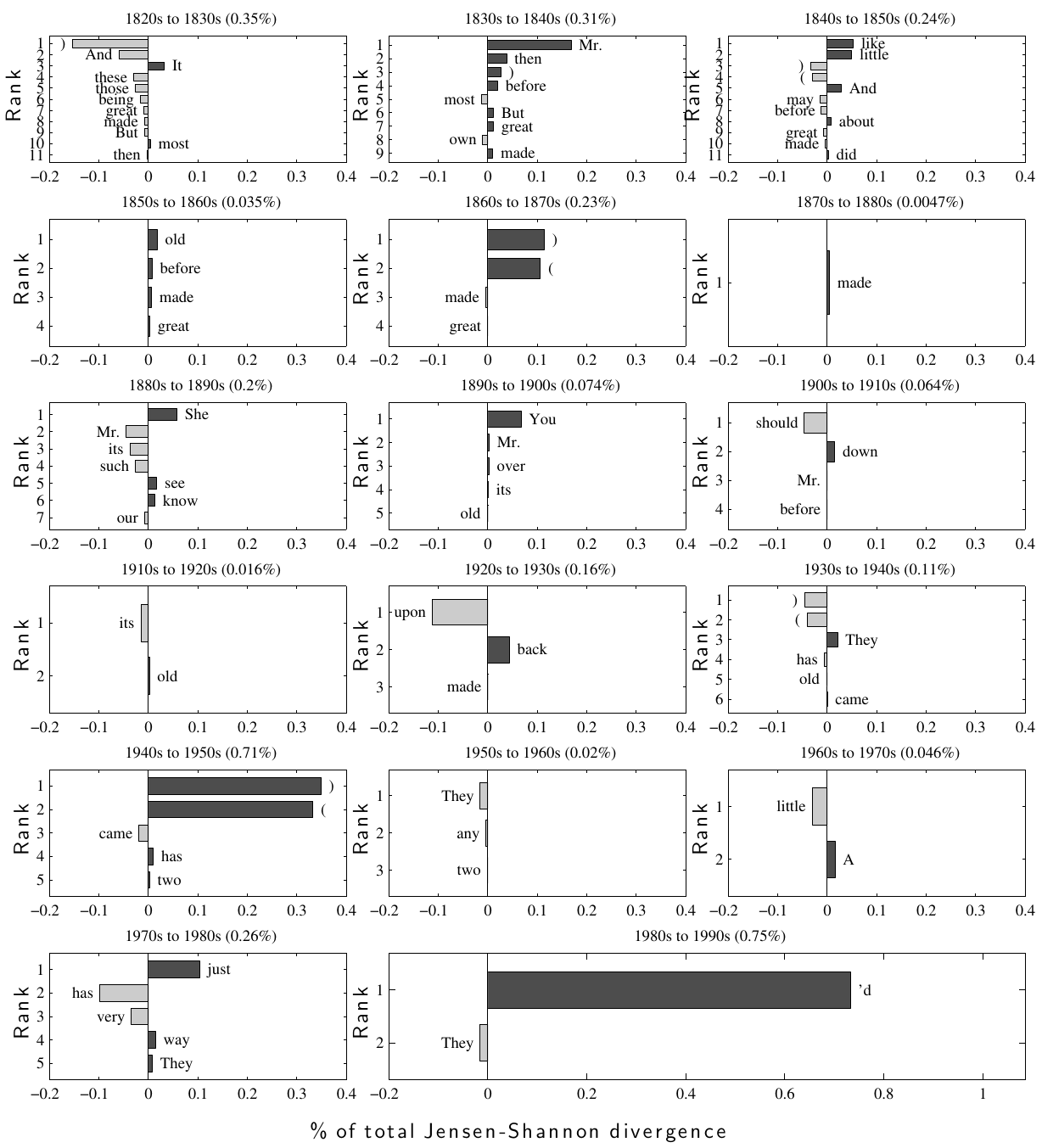}
    \caption{ 
      Words crossing relative frequency threshold of $10^{-3}$
      between consecutive decades. Signals for each pair of decades
      are sorted and weighted by contribution to the
      Jensen-Shannon divergence
      (JSD) between those
      decades. Bars pointing to the right represent words that rose
      above the threshold between decades. Bars pointing left
      represent words that fell. In parentheses in each title is the
      total percent of the JSD between the given pair of decades that
      is accounted for by flux over the $10^{-3}$ threshold.
    }
    \label{fig:langevo.threshold_flux_3}
\end{figure*}

\begin{figure}[tbp!]
  \centering
  \includegraphics[width=0.45\textwidth]{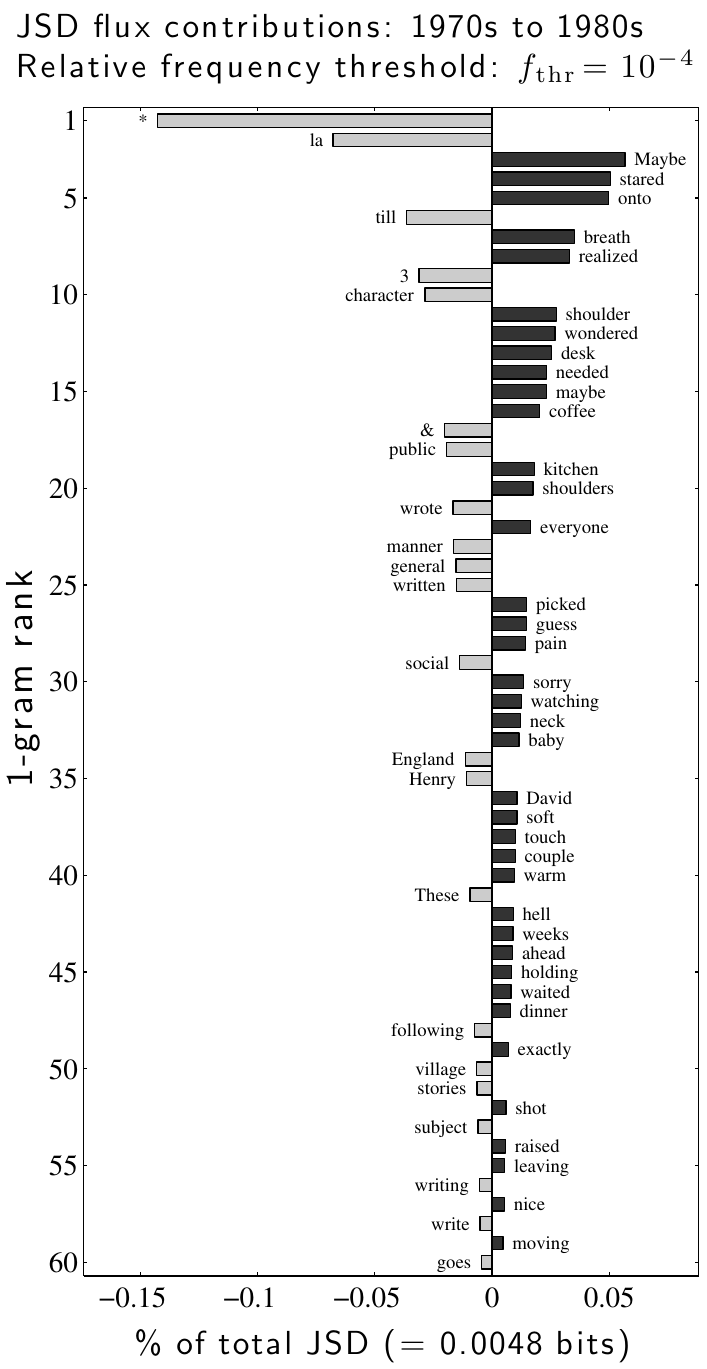}
  \caption{ 
    The top 1-grams crossing relative frequency threshold of $10^{-4}$
    between the 1970s and 1980s.
    Words are rank-ordered by their contribution to the
    Jensen-Shannon divergence (JSD) between those
    decades.
    Bars pointing to the right represent 1-grams that rose
    above the threshold between decades, and bars pointing left represent
    1-grams that fell. (The first signal is the asterisk ``*''.)
    We omit references to years in this and all subsequent figures.
  }
  \label{fig:langevo.threshold_flux_4_70}
\end{figure}

\begin{figure}[tbp!]
  \centering
  \includegraphics[width=0.45\textwidth]{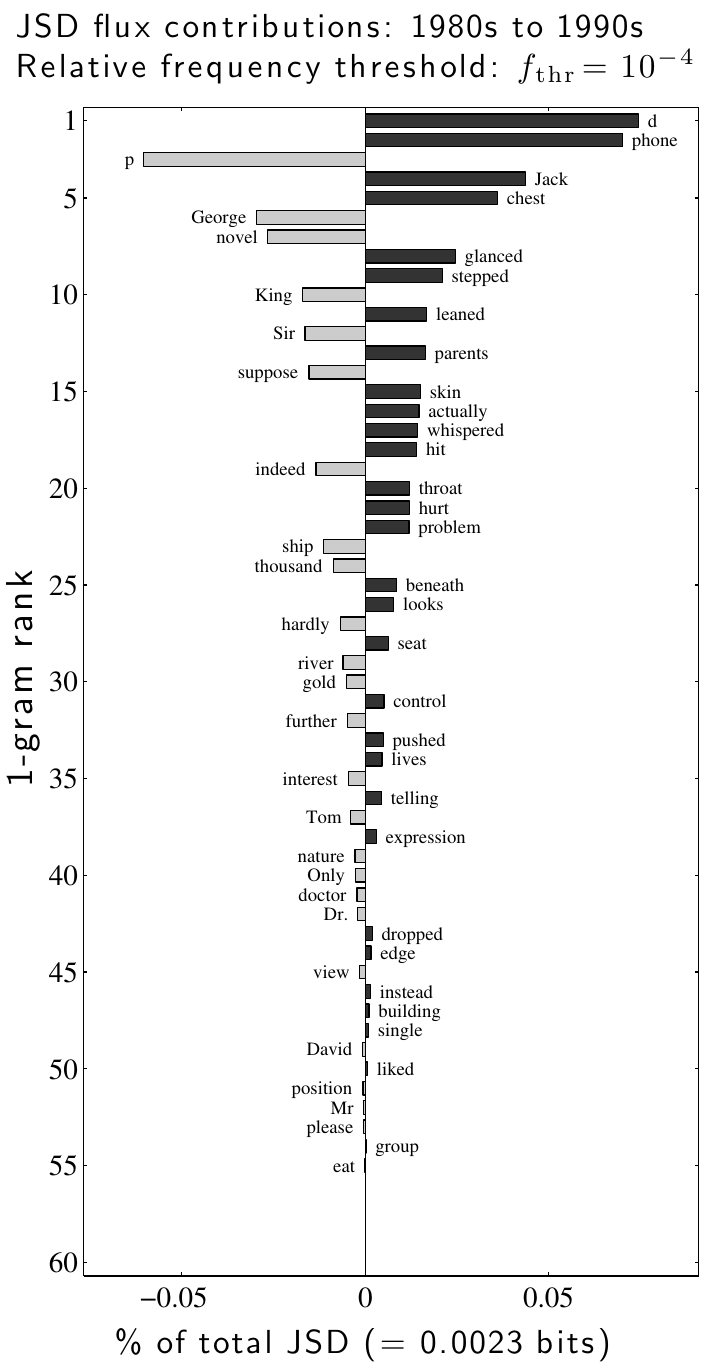}
  \caption{
    Words (not counting references to years) crossing relative frequency threshold of $10^{-4}$
    between the 1980s and 1990s.
    See the caption for Fig.~\ref{fig:langevo.threshold_flux_4_70} for details.
    Note that only 55 words make this transition.
  }
  \label{fig:langevo.threshold_flux_4_80}
\end{figure}

\begin{figure}[tbp!]
  \centering
  \includegraphics[width=0.45\textwidth]{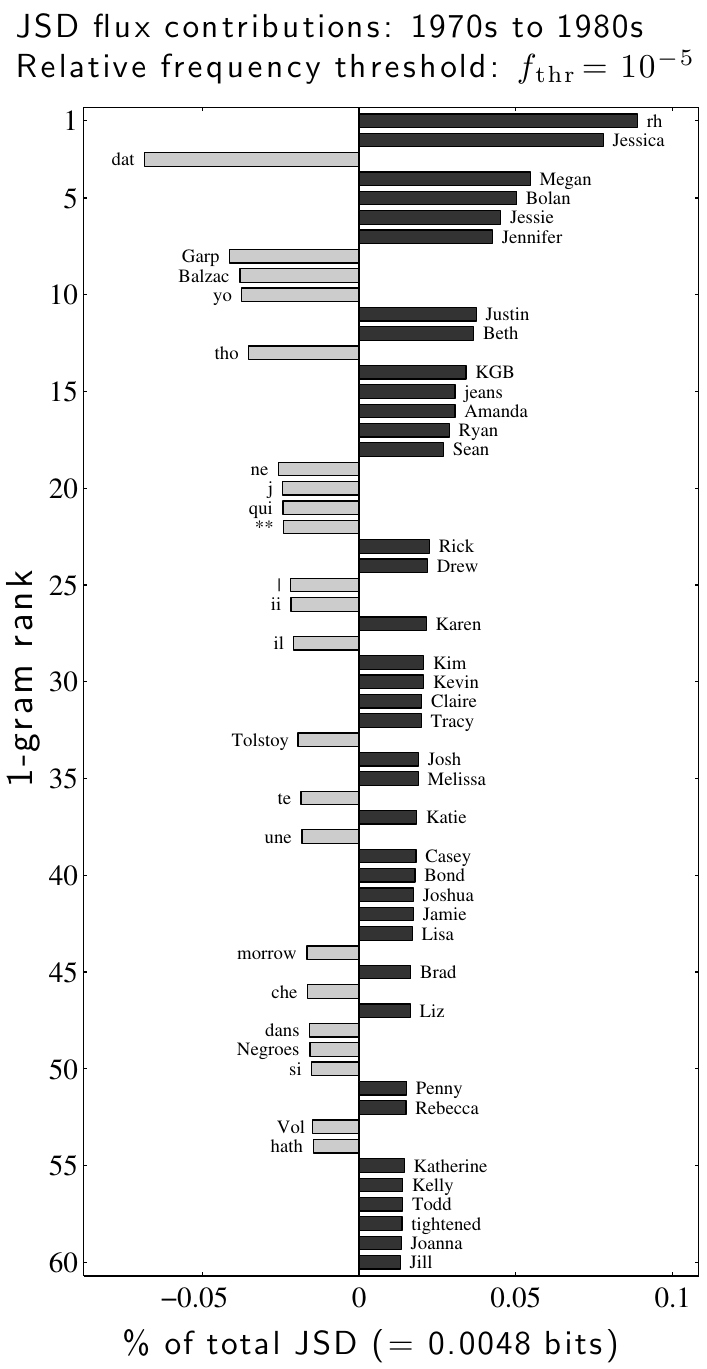}
  \caption{ 
    Words (not counting references to years) crossing relative
    frequency threshold of $10^{-5}$ between 
    the 1970s and 1980s.
    See the caption for Fig.~\ref{fig:langevo.threshold_flux_4_70} for details.
  }
  \label{fig:langevo.threshold_flux_5_70}
\end{figure}

\begin{figure}[tbp!]
  \centering
  \includegraphics[width=0.45\textwidth]{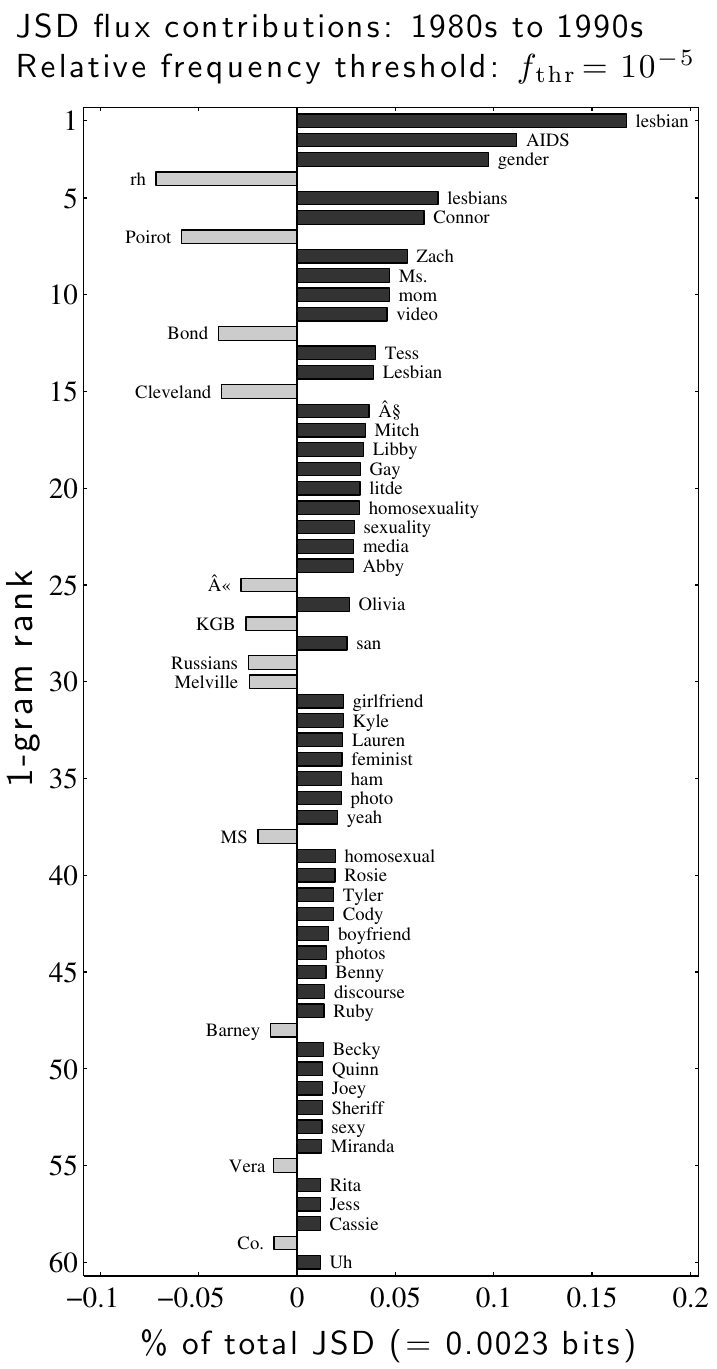}
  \caption{
    Words (not counting references to years) crossing relative
    frequency threshold of $10^{-5}$ between the 1980s and 1990s.
    See the caption for Fig.~\ref{fig:langevo.threshold_flux_4_70} for details.
  }
  \label{fig:langevo.threshold_flux_5_80}
\end{figure}

\begin{figure}[tbp!]
  \centering
  \includegraphics[width=0.45\textwidth]{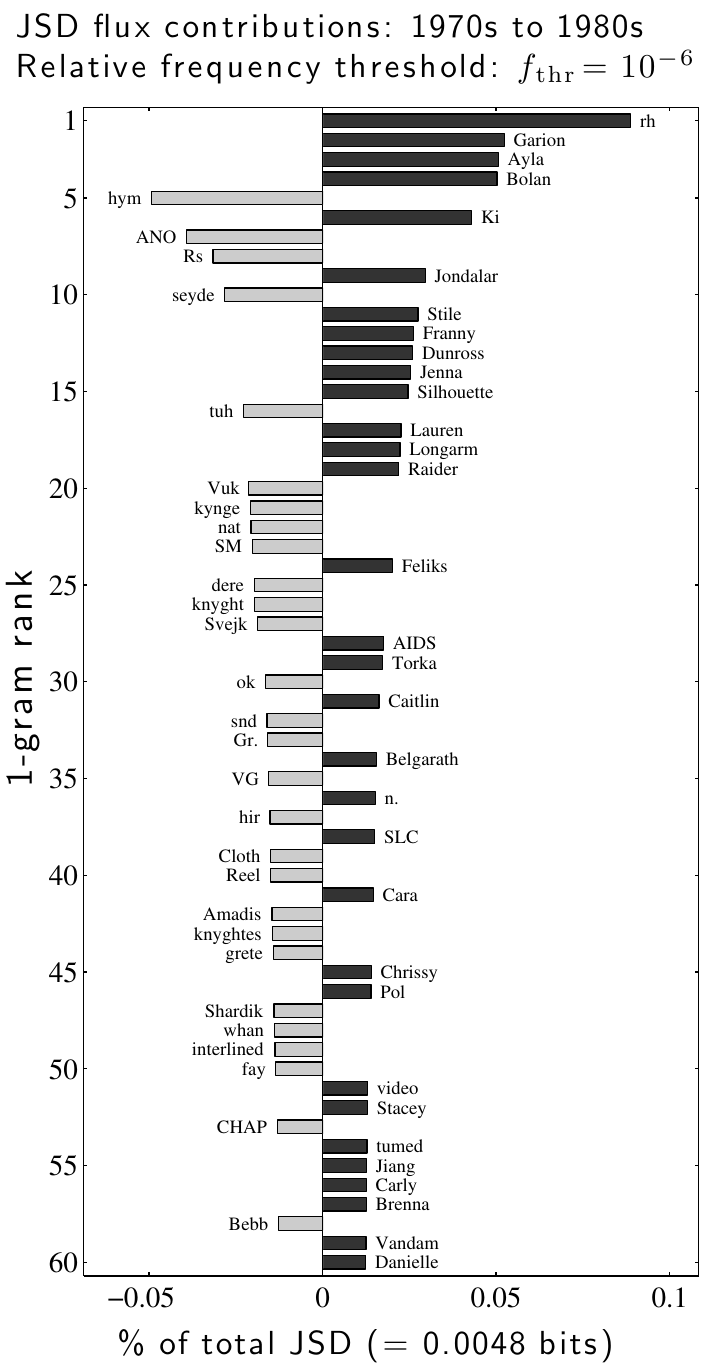}
  \caption{
    Words (not counting references to years) crossing relative
    frequency threshold of $10^{-6}$ between the 1970s and 1980s.
    See the caption for Fig.~\ref{fig:langevo.threshold_flux_4_70} for details.
  }
  \label{fig:langevo.threshold_flux_6_70}
\end{figure}

\begin{figure}[tbp!]
  \centering
  \includegraphics[width=0.45\textwidth]{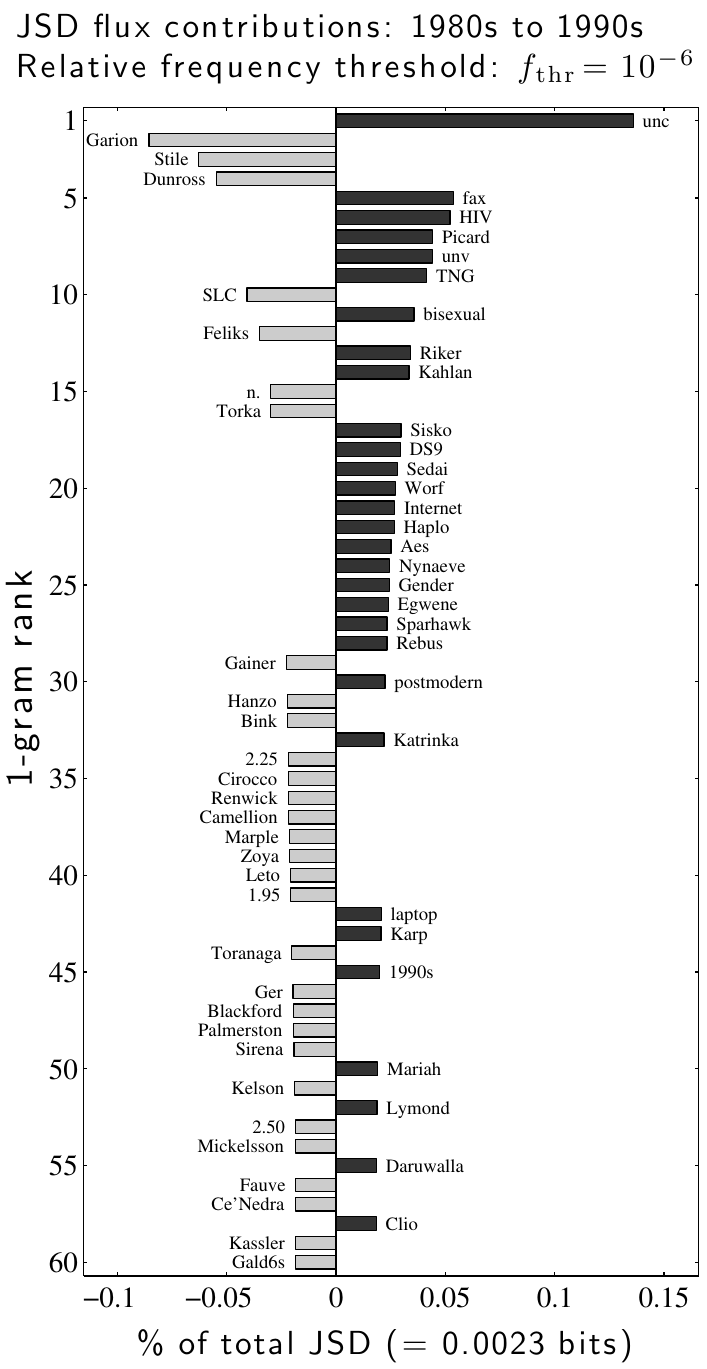}
  \caption{
    Words (not counting references to years) crossing relative
    frequency threshold of $10^{-6}$ between the 1980s and 1990s.
    See the caption for Fig.~\ref{fig:langevo.threshold_flux_4_70} for details.
  }
  \label{fig:langevo.threshold_flux_6_80}
\end{figure}

\begin{figure}[tbp!]
  \centering
  \includegraphics[width=0.45\textwidth]{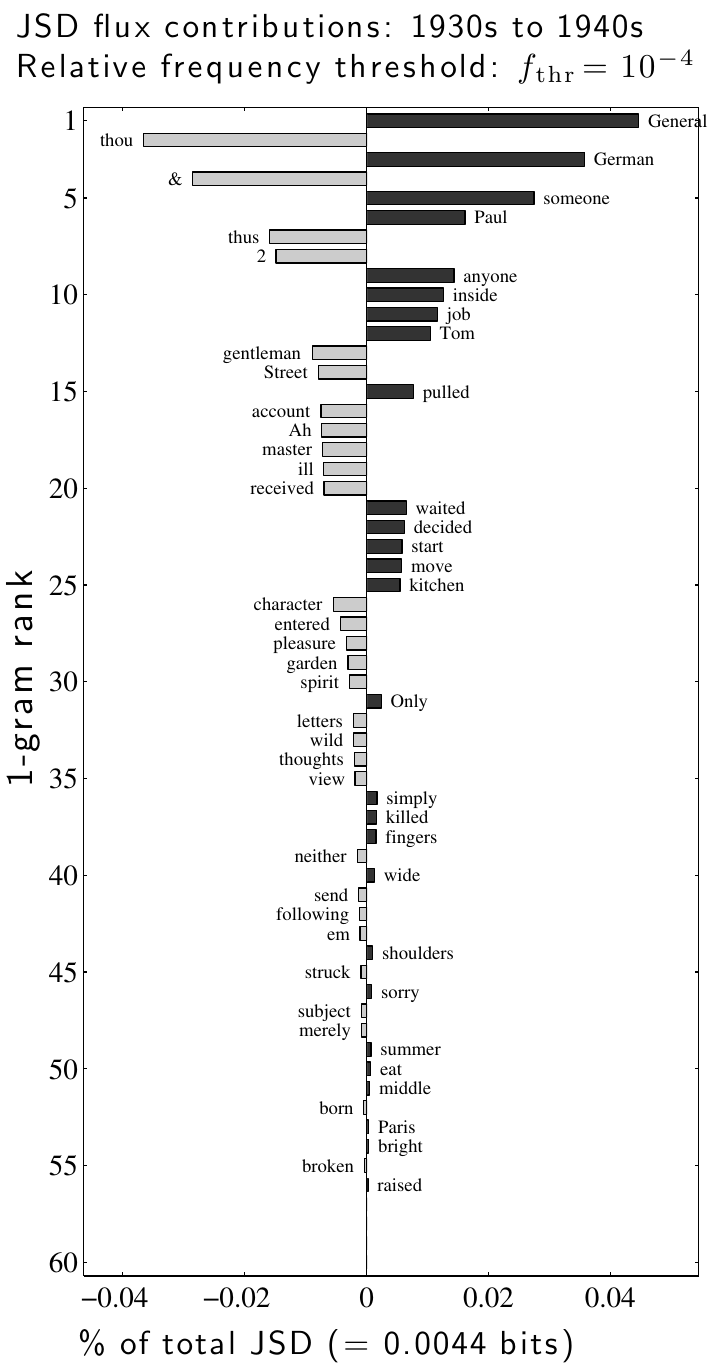}
  \caption{
    Words (not counting references to years) crossing relative
    frequency threshold of $10^{-4}$
    between the 1930s and 1940s.
    See the caption for Fig.~\ref{fig:langevo.threshold_flux_4_70} for details.
    Note that only 56 words make this transition.
  }
  \label{fig:langevo.threshold_flux_4_30}
\end{figure}

\begin{figure}[tbp!]
  \centering
  \includegraphics[width=0.45\textwidth]{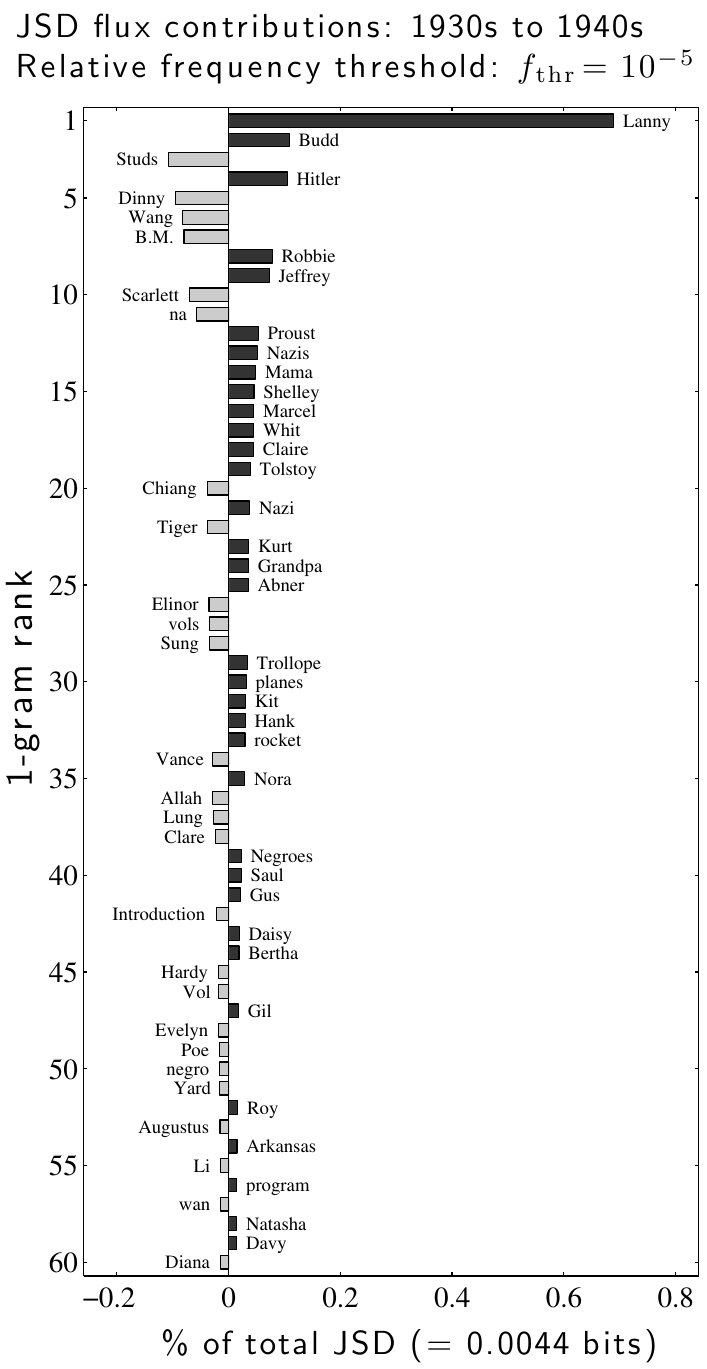}
  \caption{
    Words (not counting references to years) crossing relative
    frequency threshold of $10^{-5}$ between the 1930s and 1950s.
    See the caption for Fig.~\ref{fig:langevo.threshold_flux_4_70} for details.
  }
  \label{fig:langevo.threshold_flux_5_30}
\end{figure}

\begin{figure}[tbp!]
  \centering
  \includegraphics[width=0.45\textwidth]{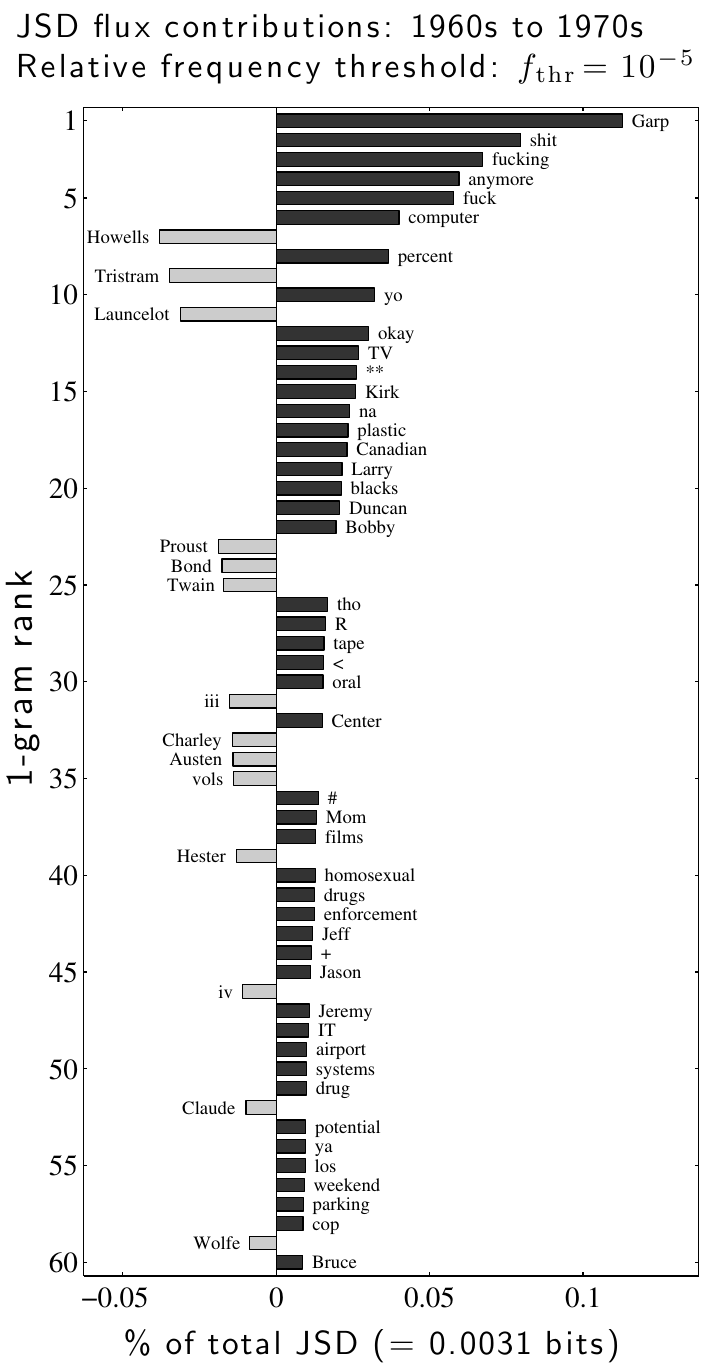}
  \caption{
    Words (not counting references to years) crossing relative
    frequency threshold of $10^{-5}$ between the 1960s and 1970s.
    See the caption for Fig.~\ref{fig:langevo.threshold_flux_4_70} for details.
  }
  \label{fig:langevo.threshold_flux_5_60}
\end{figure}

The set of 1-grams with relative frequencies above $10^{-3}$
is also fairly stable.
From Zipf's law for the 2012 English Fiction corpus
(Fig~\ref{fig:langevo.zipfslaw}), we know that this threshold
is typically exceeded by the 100 most common 1-grams.
Viewing language as code, these top 100 1-grams are fundamental
elements in the construction of meaningful statements
and comprise around 55\% of all 1-grams.

We should expect limited turnover for these core 1-grams,
and indeed the flux of 1-grams across the $10^{-3}$ boundary between consecutive decades
is entirely captured by Fig.~\ref{fig:langevo.threshold_flux_3}.

We generate these and all subsequent ``JSD word shift'' figures by:
\begin{enumerate}
\item 
  Finding all 1-grams that either move above or below a given threshold between
  two decades;
\item 
  Ranking these 1-grams by their  contributions to the JSD measured between the same decades;
  and
\item 
  Plotting downward flux 1-grams with their contributions as bars to the left,
  and upward flux 1-grams with their contributions as bars to the right.
  We leave aside all 1-grams representing years.
\end{enumerate}

In taking a close look at Fig.~\ref{fig:langevo.threshold_flux_3},
we see that parentheses drop
in relative frequency of use between the 1840s and 1850s and cross
back over the threshold after the American Civil War (between the
1860s and 1870s).
The same is true for before and after World War II (between the 1930s
and 1940s and between the 1940s and 1950s, respectively).
Beyond these, the flux is entirely due to proper words (not
punctuation).
For example, ``made'' fluctuates up and down over this threshold
repeatedly over the course of a century.
Between the 1870s and the 1880s, ``made'', which sees slightly
increased use, is the only word to cross the threshold.
The most crossings is 12, which occurs between the first two decades.
Also, ``great'' struggled over the first 5 decades and eventually
failed to remain great by this measure.
``Mr.'' fluctuated across the
threshold between the 1830s and 1910s. More recently,
from the 1930s on,
``They'' has been making its paces up and down across the
threshold.

\subsection{Fine-grained exploration of flux across frequency threshold boundaries: 1970s--1980s and 1980s--1990s}
\label{langevo:subsec.detailedflux10-1970s1980s1990s}

We now choose a few interesting decade-to-decade transitions
to delve into for the flux at the lower frequency thresholds
of
$
\frequencythreshold
=
10^{-4}$,
$10^{-5}$,
and
$10^{-6}$.
Returning to Fig.~\ref{fig:langevo.threshold_crossings},
we know that for each threshold between $10^{-4}$ and $10^{-7}$, the upward and
downward flux roughly cancel.
For both upward and downward flux, there appears to be little
qualitative difference between the three smallest thresholds
of
$10^{-4}$,
$10^{-5}$,
and
$10^{-6}$.
For these thresholds,
the downward flux between the 1950s and the 1960s is a minimum
then increases over the next two pairs of
consecutive decades,
and then dips again between the 1980s and
1990s.
For $\frequencythreshold=10^{-4}$,
the increase between the 1960s and
1970s and the next pair of decades is more noticeable for the downward
flux, as is the decrease between the last two pairs of decades.

Based on these observations, we will examine the flux for two decade pairs
in this section: 1970s--1980s and 1980s--1990s.
In the following two sections, we will consider the 1930s--1940s
transition because of the historical importance of these decades,
and then finally the 1960s--1970s transition to show some peculiarities
of word flux.

We begin
by displaying in Fig.~\ref{fig:langevo.threshold_flux_4_70} the top 60 flux 1-grams
across $\frequencythreshold$=$10^{-4}$
between the 1970s and the 1980s, and
in Fig.~\ref{fig:langevo.threshold_flux_4_80},
we show all 55 flux words between the 1980s and the 1990s
for the same $\frequencythreshold$.
In these and all subsequent figures,
we use the same format as
Fig.~\ref{fig:langevo.threshold_flux_3}.

Between each pair of decades, we see reduced relative use of
particularly British words, including ``England'' between the first two
decades and ``King'', ``George'', and ``Sir'' between the latter
two. We also see reduced use of more formal-sounding words, such as
``character'', ``manner'', and ``general'' between the first two
decades and ``suppose'', ``indeed'', and ``hardly'' between the latter
two. Increasing are physical and emotional words. Those between the
first two decades include ``stared'', ``breath'', ``realized'',
``shoulder'' and ``shoulders'', ``coffee'', ``guess'', ``pain'', and
``sorry.'' Between the latter two, we see ``chest'', ``skin'',
``whispered'', ``hit'', ``throat'', ``hurt'', ``control'', and
``lives.'' Also included are ``phone'' and ``parents.''

In Figs.~\ref{fig:langevo.threshold_flux_5_70}~and~\ref{fig:langevo.threshold_flux_5_80},
we display the top 60 flux words, not counting references to years,
across the $10^{-5}$ threshold between the same pairs of decades.
Many of the words declining below the threshold between the 1970s and 1980s are
unusual spellings such as ``tho'', proper names like ``Balzac'', or
words from non-English languages like ``une.''
Increasing across this threshold between the first
two decades are a plethora of mostly female proper names, with
``Jessica'' and ``Megan'' leading. Also seen are ``KGB'' and
``jeans.'' (``KGB'' decreases in the 1990s, as does ``Russians.'')
Increasing between the 1980s and 1990s are a few proper names;
however, most of the signals here are social and sexual in
nature, and in part point to the inclusion of academic, literary criticism.
These include ``lesbian'' and ``lesbians'', ``AIDS'', and
``gender'' in the top positions. Also included are both
``homosexuality'' and the more general ``sexuality.'' We also see
``girlfriend'', ``boyfriend'', ``feminist'', and ``sexy.''

We show in Figs.~\ref{fig:langevo.threshold_flux_6_70}
and~\ref{fig:langevo.threshold_flux_6_80}
the flux across a threshold of $10^{-6}$ between the 1970s and
1980s, and the 1980s and 1990s
(again, not counting years).
The first of these is not particularly topical, though we do
see ``AIDS'' increase above this threshold a decade prior to its
increase over $10^{-5}$ as seen in Fig.~\ref{fig:langevo.threshold_flux_5_80}.
For the second pair of decades, we find some surprising signals.
In particular, while increases in ``HIV''
and ``bisexual'' make the list (similarly to many signals in
Fig.~\ref{fig:langevo.threshold_flux_5_80}), as do ``fax'', ``laptop'', and
``Internet'', a great swath of the signals are accounted for by one
franchise.
We note increases in ``Picard'', ``TNG'', ``Sisko'', and
``DS9.''
These latter signals should serve as a reminder that the
word distributions in library-like Google Books
corpus~\cite{pechenick2015a}, 
even for fiction, do not remotely resemble the contents of 
normal conversations (at least not for the general population).
However, we do observe signals arising at this
threshold from factors external to the imaginings of specific authors.
It would therefore be premature to dismiss the contributions at this
threshold because of an apparent overabundance of ``Star Trek.'' In
fact, because ``The Next Generation'' and ``Deep Space 9'' aired
precisely during these two decades, an abundance of ``Star Trek''
novels in the English Fiction data set is actually quite encouraging,
because these novels do exist, are available in English, and are
(clearly) fiction.

The cultural signals change as we dial down the frequency threshold.
We typically find that thresholds of $10^{-4}$ and above
produce signals with little to no noise.
This is not surprising because
this relative frequency roughly corresponds to rank threshold for the
1000 most common words (see Fig.~\ref{fig:langevo.zipfslaw}) in the data
set. Using a threshold of $10^{-5}$ (fewer than 10,000 words fall
above this frequency in any given decade), we see some noise (mostly
in the form of familiar names), but still observe many valuable
signals. Only when the threshold is reduced to $10^{-6}$ does the
overall texture of the signals become questionable as a result of a
variety of proper nouns far less familiar than those observed with the
previous threshold. However, at this threshold, we nevertheless observe
several early signals of real social importance.

\subsection{Fine-grained exploration of flux across frequency threshold boundaries: 1930s--1940s}
\label{langevo:subsec.detailedflux10-1930s1940s}

Curiously, between the 1930s and 1940s the volume of flux across each
threshold is not atypical (see
Fig.~\ref{fig:langevo.threshold_crossings}). Moreover, the asymmetry between the
JSD contributions between those decades is very low. Yet it is obvious
that we should expect signals of historical significance between these
two decades, and indeed we do once we examine the dynamics of individual 1-grams.
In
Figs.~\ref{fig:langevo.threshold_flux_4_30}~and~\ref{fig:langevo.threshold_flux_5_30}, we see
words crossing the $10^{-4}$ and $10^{-5}$ thresholds, respectively
(with references to years omitted in
Fig.~\ref{fig:langevo.threshold_flux_5_30}). For the higher threshold, only 56
words cross. The most noticeable such words that are more commonly
used in the 1940s are ``General'' and ``German.'' Also, ``killed''
appears in this list. Words used less frequently include ``pleasure'',
``garden'', and ``spirit.'' For the lower threshold, we see the
signals from prolific authors as in our previous
paper~\cite{pechenick2015a}, particularly Upton
Sinclair's character, Lanny Budd. We also see more Nazis (``Nazi'' and
``Nazis'').

\subsection{Fine-grained exploration of flux across frequency threshold boundaries: 1960s--1970s}
\label{langevo:subsec.detailedflux10-1960s1970s}

Last, we include one of the more colorful examples. In
Fig.~\ref{fig:langevo.threshold_flux_5_60}, we show signals (not including years)
for words crossing the $10^{-5}$ threshold between the 1960s and
1970s. Profanity dominates.
We see references to \textit{The
  World According to Garp} (``Garp'') and, again, to ``Star Trek''
(``Kirk'' this time).
We also see more ``computer'', ``TV'', and, per The Graduate, ``plastic.''
Signals also appear for ``blacks'' and ``homosexual'', for narcotics
(``drug'' and ``drugs''), and a changing role for police (``enforcement'' and ``cop'').

We refer the reader to our paper's Online Appendices~\cite{pechenick2015b_onlineappendices}
for figures representing flux across relative frequency thresholds of
$10^{-4}$, $10^{-5}$, and $10^{-6}$ between consecutive decades over
the entire period analyzed (the 1820s to the 1990s).

\section{Concluding remarks}
\label{sec:langevo.conc}

In seeking to characterize word birth and death for the 2012 Google Books
English Fiction corpus,
we have identified and characterized what we believe is
a fundamental feature of
language evolution: lexical turbulence.
In general, for any time-coded corpus,
we quantify lexical turbulence as the flux $\flux$ of words across a relative
frequency of usage threshold between two time periods.
We speak of undirected flux $\flux$ because
we found that upward and downward flux
$\fluxup$
and
$\fluxdown$
across a threshold
were on average well balance,
though this may not always be the case.

Like the Jensen-Shannon divergence and related measures,
lexical turbulence is one way of characterizing
the degree of word rank (or relative frequency)
variability underlying Zipf's law.
The overall form of Zipf's law may be strongly preserved
across corpora suggesting stability
(Fig.~\ref{fig:langevo.zipfslaw})
but completely occlude
how individual word usage rates are changing
(Fig.~\ref{fig:langevo.threshold_crossings}).

Word flux may also be naturally measured across a fixed word rank,
with the connection to relative frequency being made through Zipf's law
(Fig.~\ref{fig:langevo.threshold_crossings}D).
The scaling of word flux with rank is superlinear
with a break in scaling tied to that of Zipf's law~\cite{williams2015b}.

We conjecture that word rank may be viewed as roughly analogous to a kind of temperature
where the most common words are nearly frozen in usage rates
while rarer and rarer words increasingly boil and bubble in their
relative frequencies.
One metaphor for words sometimes invoked is that of tools~\cite{zipf1949a}.
Words form a hierarchy of tools with a
crystallized set of the most frequently used instruments (comma, period, ``the'')
resting above a vast tool set of increasingly specific uses.

We arrived at the notion of lexical turbulence via
our efforts to reproduce the results of~\cite{petersen2012a}.
We found general agreement regarding a decay in word birth from 1800 to 2000
but not so for word death.
True word death appears to be extremely and durably rare.
Overall, the lowering birth rate signals
a cooling of language~\cite{petersen2012b}
but the time-independent scaling of lexical turbulence
shows that the lexicon is constantly turning over.

Using JSD word shifts, we also explored in detail the words
dominating the flux across some example frequency thresholds
for a number of interesting decade-decades transitions.
While extremely specific fiction can be of great
interest---whether it be in the form of war novels or volumes from the
``Star Trek'' franchise---vocabulary from these works is more easily
studied when placed in proper context.
Dialing down the relative
frequency threshold across several orders of magnitude helps to
capture this distinction. However, further experimentation is called for,
because an automatic means of separating specific signals from the more
general signals (e.g., ``Star Trek'' from social movements) could
afford both a more intuitive grasp of the lexical dynamics and
might, ideally, allow investigators to hypothesize causal
relationships between exogenous and endogenous drivers of
language.

Of many potential directions for future work, several that stand out would be
(1) Reproducing the present analysis of lexical turbulence for
2-grams and 3-grams which, $n$-grams that are particularly rich in meaning;
(2) Quantifying the behavior of lexical turbulence with time (e.g., beyond
adjacent decade comparisons as we have done here);
(3) Creating toy models of language evolution to attempt to capture lexical turbulence;
and
(4) Building interactive JSD-based word shifts where corpora, frequency thresholds
and year range may be selected to facilitate rapid explorations.

\acknowledgments
We thank Simon DeDeo for helpful discussions.
We were able to improve our paper per excellent suggestions
from an anonymous referee.
PSD was supported by NSF CAREER Award \# 0846668.


\begin{thebibliography}{19}
\expandafter\ifx\csname natexlab\endcsname\relax\def\natexlab#1{#1}\fi
\expandafter\ifx\csname bibnamefont\endcsname\relax
  \def\bibnamefont#1{#1}\fi
\expandafter\ifx\csname bibfnamefont\endcsname\relax
  \def\bibfnamefont#1{#1}\fi
\expandafter\ifx\csname citenamefont\endcsname\relax
  \def\citenamefont#1{#1}\fi
\expandafter\ifx\csname url\endcsname\relax
  \def\url#1{\texttt{#1}}\fi
\expandafter\ifx\csname urlprefix\endcsname\relax\def\urlprefix{URL }\fi
\providecommand{\bibinfo}[2]{#2}
\providecommand{\eprint}[2][]{\url{#2}}

\bibitem[{\citenamefont{Petersen
  et~al.}(2012{\natexlab{a}})\citenamefont{Petersen, Tenenbaum, Havlin, and
  Stanley}}]{petersen2012a}
\bibinfo{author}{\bibfnamefont{A.~M.} \bibnamefont{Petersen}},
  \bibinfo{author}{\bibfnamefont{J.}~\bibnamefont{Tenenbaum}},
  \bibinfo{author}{\bibfnamefont{S.}~\bibnamefont{Havlin}}, \bibnamefont{and}
  \bibinfo{author}{\bibfnamefont{H.~E.} \bibnamefont{Stanley}},
  \bibinfo{journal}{Scientific Reports} \textbf{\bibinfo{volume}{2}},
  \bibinfo{pages}{313} (\bibinfo{year}{2012}{\natexlab{a}}).

\bibitem[{\citenamefont{Petersen
  et~al.}(2012{\natexlab{b}})\citenamefont{Petersen, Tenenbaum, Havlin,
  Stanley, and Perc}}]{petersen2012b}
\bibinfo{author}{\bibfnamefont{A.~M.} \bibnamefont{Petersen}},
  \bibinfo{author}{\bibfnamefont{J.~N.} \bibnamefont{Tenenbaum}},
  \bibinfo{author}{\bibfnamefont{S.}~\bibnamefont{Havlin}},
  \bibinfo{author}{\bibfnamefont{H.~E.} \bibnamefont{Stanley}},
  \bibnamefont{and} \bibinfo{author}{\bibfnamefont{M.}~\bibnamefont{Perc}},
  \bibinfo{journal}{Scientific reports} \textbf{\bibinfo{volume}{2}},
  \bibinfo{pages}{943} (\bibinfo{year}{2012}{\natexlab{b}}).

\bibitem[{\citenamefont{Abrams and Strogatz}(2003)}]{abrams2003a}
\bibinfo{author}{\bibfnamefont{D.~M.} \bibnamefont{Abrams}} \bibnamefont{and}
  \bibinfo{author}{\bibfnamefont{S.~H.} \bibnamefont{Strogatz}},
  \bibinfo{journal}{Nature} \textbf{\bibinfo{volume}{424}},
  \bibinfo{pages}{900} (\bibinfo{year}{2003}).

\bibitem[{\citenamefont{Zipf}(1949)}]{zipf1949a}
\bibinfo{author}{\bibfnamefont{G.~K.} \bibnamefont{Zipf}},
  \emph{\bibinfo{title}{Human Behaviour and the Principle of Least-Effort}}
  (\bibinfo{publisher}{Addison-Wesley}, \bibinfo{address}{Cambridge, MA},
  \bibinfo{year}{1949}).

\bibitem[{\citenamefont{Michel et~al.}(2011)\citenamefont{Michel, Shen, Aiden,
  Veres, Gray, Pickett, Hoiberg, Clancy, Norvig, Orwant
  et~al.}}]{michel2011quantitative}
\bibinfo{author}{\bibfnamefont{J.-B.} \bibnamefont{Michel}},
  \bibinfo{author}{\bibfnamefont{Y.~K.} \bibnamefont{Shen}},
  \bibinfo{author}{\bibfnamefont{A.~P.} \bibnamefont{Aiden}},
  \bibinfo{author}{\bibfnamefont{A.}~\bibnamefont{Veres}},
  \bibinfo{author}{\bibfnamefont{M.~K.} \bibnamefont{Gray}},
  \bibinfo{author}{\bibfnamefont{J.~P.} \bibnamefont{Pickett}},
  \bibinfo{author}{\bibfnamefont{D.}~\bibnamefont{Hoiberg}},
  \bibinfo{author}{\bibfnamefont{D.}~\bibnamefont{Clancy}},
  \bibinfo{author}{\bibfnamefont{P.}~\bibnamefont{Norvig}},
  \bibinfo{author}{\bibfnamefont{J.}~\bibnamefont{Orwant}},
  \bibnamefont{et~al.}, \bibinfo{journal}{science}
  \textbf{\bibinfo{volume}{331}}, \bibinfo{pages}{176} (\bibinfo{year}{2011}).

\bibitem[{\citenamefont{Lin et~al.}(2012)\citenamefont{Lin, Michel, Aiden,
  Orwant, Brockman, and Petrov}}]{lin2012syntactic}
\bibinfo{author}{\bibfnamefont{Y.}~\bibnamefont{Lin}},
  \bibinfo{author}{\bibfnamefont{J.-B.} \bibnamefont{Michel}},
  \bibinfo{author}{\bibfnamefont{E.~L.} \bibnamefont{Aiden}},
  \bibinfo{author}{\bibfnamefont{J.}~\bibnamefont{Orwant}},
  \bibinfo{author}{\bibfnamefont{W.}~\bibnamefont{Brockman}}, \bibnamefont{and}
  \bibinfo{author}{\bibfnamefont{S.}~\bibnamefont{Petrov}}, in
  \emph{\bibinfo{booktitle}{Proceedings of the ACL 2012 System Demonstrations}}
  (\bibinfo{organization}{Association for Computational Linguistics},
  \bibinfo{year}{2012}), pp. \bibinfo{pages}{169--174}.

\bibitem[{\citenamefont{Pechenick et~al.}(2015)\citenamefont{Pechenick,
  Danforth, and Dodds}}]{pechenick2015a}
\bibinfo{author}{\bibfnamefont{E.~A.} \bibnamefont{Pechenick}},
  \bibinfo{author}{\bibfnamefont{C.~M.} \bibnamefont{Danforth}},
  \bibnamefont{and} \bibinfo{author}{\bibfnamefont{P.~S.} \bibnamefont{Dodds}},
  \bibinfo{journal}{PLoS ONE} \textbf{\bibinfo{volume}{10}},
  \bibinfo{pages}{e0137041} (\bibinfo{year}{2015}).

\bibitem[{\citenamefont{Gerlach and
  Altmann}(2013{\natexlab{a}})}]{gerlach2013stochastic}
\bibinfo{author}{\bibfnamefont{M.}~\bibnamefont{Gerlach}} \bibnamefont{and}
  \bibinfo{author}{\bibfnamefont{E.~G.} \bibnamefont{Altmann}},
  \bibinfo{journal}{Physical Review X} \textbf{\bibinfo{volume}{3}},
  \bibinfo{pages}{021006} (\bibinfo{year}{2013}{\natexlab{a}}).

\bibitem[{pec()}]{pechenick2015b_onlineappendices}
\bibinfo{note}{Online Appendices can be found at:
  \url{http://compstorylab.org/share/papers/pechenick2015b/}}.

\bibitem[{\citenamefont{Pratchett}(1997)}]{pratchett1997b}
\bibinfo{author}{\bibfnamefont{T.}~\bibnamefont{Pratchett}},
  \emph{\bibinfo{title}{Feet of Clay}} (\bibinfo{publisher}{HarperCollins},
  \bibinfo{address}{New York}, \bibinfo{year}{1997}).

\bibitem[{mon(1969)}]{montypython1969deadparrotsketch}
\emph{\bibinfo{title}{Dead parrot sketch}} (\bibinfo{year}{1969}),
  \bibinfo{note}{{M}onty Python's Flying Circus;
  \url{https://en.wikipedia.org/wiki/Dead_Parrot_sketch}}.

\bibitem[{\citenamefont{Williams et~al.}(2015)\citenamefont{Williams, Bagrow,
  Danforth, and Dodds}}]{williams2015b}
\bibinfo{author}{\bibfnamefont{J.~R.} \bibnamefont{Williams}},
  \bibinfo{author}{\bibfnamefont{J.~P.} \bibnamefont{Bagrow}},
  \bibinfo{author}{\bibfnamefont{C.~M.} \bibnamefont{Danforth}},
  \bibnamefont{and} \bibinfo{author}{\bibfnamefont{P.~S.} \bibnamefont{Dodds}},
  \bibinfo{journal}{Physical Review E} \textbf{\bibinfo{volume}{91}},
  \bibinfo{pages}{052811} (\bibinfo{year}{2015}).

\bibitem[{\citenamefont{Ferrer-i Cancho and
  Sol\'{e}}(2001)}]{ferrericancho2001c}
\bibinfo{author}{\bibfnamefont{R.}~\bibnamefont{Ferrer-i Cancho}}
  \bibnamefont{and} \bibinfo{author}{\bibfnamefont{R.~V.}
  \bibnamefont{Sol\'{e}}}, \bibinfo{journal}{Journal of Quantitative
  Linguistics} \textbf{\bibinfo{volume}{8}}, \bibinfo{pages}{165}
  (\bibinfo{year}{2001}).

\bibitem[{\citenamefont{Gerlach and
  Altmann}(2013{\natexlab{b}})}]{gerlach2013a}
\bibinfo{author}{\bibfnamefont{M.}~\bibnamefont{Gerlach}} \bibnamefont{and}
  \bibinfo{author}{\bibfnamefont{E.~G.} \bibnamefont{Altmann}},
  \bibinfo{journal}{Phys. Rev. X} \textbf{\bibinfo{volume}{3}},
  \bibinfo{pages}{021006} (\bibinfo{year}{2013}{\natexlab{b}}).

\bibitem[{\citenamefont{Lin}(1991)}]{lin1991divergence}
\bibinfo{author}{\bibfnamefont{J.}~\bibnamefont{Lin}},
  \bibinfo{journal}{Information Theory, IEEE Transactions on}
  \textbf{\bibinfo{volume}{37}}, \bibinfo{pages}{145} (\bibinfo{year}{1991}).

\bibitem[{\citenamefont{Klingenstein et~al.}(2014)\citenamefont{Klingenstein,
  Hitchcock, and DeDeo}}]{klingenstein2014a}
\bibinfo{author}{\bibfnamefont{S.}~\bibnamefont{Klingenstein}},
  \bibinfo{author}{\bibfnamefont{T.}~\bibnamefont{Hitchcock}},
  \bibnamefont{and} \bibinfo{author}{\bibfnamefont{S.}~\bibnamefont{DeDeo}},
  \bibinfo{journal}{Proc. Natl. Acad. Sci.} \textbf{\bibinfo{volume}{111}},
  \bibinfo{pages}{9419} (\bibinfo{year}{2014}).

\bibitem[{\citenamefont{Shannon}(1948)}]{shannon1948a}
\bibinfo{author}{\bibfnamefont{C.~E.} \bibnamefont{Shannon}},
  \bibinfo{journal}{The Bell System Tech. J.} \textbf{\bibinfo{volume}{27}},
  \bibinfo{pages}{379} (\bibinfo{year}{1948}).

\bibitem[{\citenamefont{Dodds et~al.}(2011)\citenamefont{Dodds, Harris,
  Kloumann, Bliss, and Danforth}}]{dodds2011e}
\bibinfo{author}{\bibfnamefont{P.~S.} \bibnamefont{Dodds}},
  \bibinfo{author}{\bibfnamefont{K.~D.} \bibnamefont{Harris}},
  \bibinfo{author}{\bibfnamefont{I.~M.} \bibnamefont{Kloumann}},
  \bibinfo{author}{\bibfnamefont{C.~A.} \bibnamefont{Bliss}}, \bibnamefont{and}
  \bibinfo{author}{\bibfnamefont{C.~M.} \bibnamefont{Danforth}},
  \bibinfo{journal}{PLoS ONE} \textbf{\bibinfo{volume}{6}},
  \bibinfo{pages}{e26752} (\bibinfo{year}{2011}).

\bibitem[{\citenamefont{Dodds et~al.}(2015)\citenamefont{Dodds, Clark, Desu,
  Frank, Reagan, Williams, Mitchell, Harris, Kloumann, Bagrow
  et~al.}}]{dodds2015a}
\bibinfo{author}{\bibfnamefont{P.~S.} \bibnamefont{Dodds}},
  \bibinfo{author}{\bibfnamefont{E.~M.} \bibnamefont{Clark}},
  \bibinfo{author}{\bibfnamefont{S.}~\bibnamefont{Desu}},
  \bibinfo{author}{\bibfnamefont{M.~R.} \bibnamefont{Frank}},
  \bibinfo{author}{\bibfnamefont{A.~J.} \bibnamefont{Reagan}},
  \bibinfo{author}{\bibfnamefont{J.~R.} \bibnamefont{Williams}},
  \bibinfo{author}{\bibfnamefont{L.}~\bibnamefont{Mitchell}},
  \bibinfo{author}{\bibfnamefont{K.~D.} \bibnamefont{Harris}},
  \bibinfo{author}{\bibfnamefont{I.~M.} \bibnamefont{Kloumann}},
  \bibinfo{author}{\bibfnamefont{J.~P.} \bibnamefont{Bagrow}},
  \bibnamefont{et~al.}, \bibinfo{journal}{Proc. Natl. Acad. Sci.}
  \textbf{\bibinfo{volume}{112}}, \bibinfo{pages}{2389} (\bibinfo{year}{2015}),
  \bibinfo{note}{available online at
  \url{http://www.pnas.org/content/112/8/2389}}.

\end{thebibliography}
\end{document}